\pdfoutput=1
\documentclass{article}
\usepackage{paraweave_preprint,times}
\usepackage{iftex}
\usepackage{wrapfig}

\usepackage{amsmath,amsfonts,bm}

\def\eqref#1{equation~\ref{#1}}

\def\1{\bm{1}}

\DeclareMathAlphabet{\mathsfit}{\encodingdefault}{\sfdefault}{m}{sl}
\SetMathAlphabet{\mathsfit}{bold}{\encodingdefault}{\sfdefault}{bx}{n}

\let\ab\allowbreak

\usepackage{tcolorbox}
\usepackage[hidelinks]{hyperref}
\usepackage{url}
\usepackage{booktabs}
\usepackage{enumitem}
\usepackage{array}
\usepackage{multirow}
\usepackage{glossaries}
\usepackage{listings}
\usepackage{float}
\usepackage{microtype}
\usepackage{graphicx,tikz}
\usepackage{amsmath,amssymb}

\input{visuals}
\makeatletter
\@ifundefined{promptbox}{\newtcolorbox{promptbox}{colback=blue!2!white,colframe=black!18,boxrule=0.45pt,arc=2pt,
  left=10pt,right=10pt,top=8pt,bottom=8pt,before skip=8pt,after skip=8pt}}{}
\makeatother

\title{Finding the Parallels: A Path to Evaluating Summarization from Low-Resource Languages}

\newglossaryentry{approach}{
    name={\textsc{ParaWeave}},
    description={Method name}
}

\author{%
Ahmed Jaber\thanks{Corresponding authors: \texttt{amj2234@columbia.edu}, \texttt{kathy@cs.columbia.edu}.}\hspace{0.5em}, Zachary Horvitz, Nicholas Deas, Parmida Shariat, Kathleen McKeown\footnotemark[1] \\
Columbia University
}
\newcommand{\paperfunding}{%
\begingroup
\renewcommand{\thefootnote}{}%
\footnotetext{Our code and benchmark are available at \url{https://github.com/Ajaberr/ParaWeave}.}%
\footnotetext{This research is based upon work supported in part by the Office of the Director of National Intelligence (ODNI), Intelligence Advanced Research Projects Activity (IARPA), via 560000C260018. The views and conclusions contained herein are those of the authors and should not be interpreted as necessarily representing the official policies, either expressed or implied, of ODNI, IARPA, or the U.S. Government. The U.S. Government is authorized to reproduce and distribute reprints for governmental purposes notwithstanding any copyright annotation therein.}%
\endgroup
}
\newcommand{\paperacknowledgments}{}

\begin{document}

\maketitle
\paperfunding
\begin{abstract}
  Across benchmarks, large language models have demonstrated impressive summarization capabilities for high-resource languages. In contrast, summarization benchmarks are largely unavailable for low-resource languages. This scarcity poses a challenge for both evaluating LLM summarization capabilities and measuring the effectiveness of LLM-as-a-Judge approaches in these languages. However, previous work in machine translation has contributed parallel datasets that pair sentences in low-resource languages with higher-resource translations. We introduce \gls{approach}, a method for automatically transforming these existing sentence-level parallel datasets into coherent document-level benchmarks while preserving their original cross-lingual alignment. We used \gls{approach} to construct a summarization benchmark that spans ten language varieties from three families (Afro-Asiatic, Indo-Iranian, and Nilo-Saharan). We use the resulting benchmark to evaluate summarization faithfulness and LLM-as-a-Judge performance for summaries generated from documents in these languages. Pairing low-resource language documents with high-resource translations enables studying the effect of source language on both summarization quality and LLM-as-a-Judge performance. These comparisons reveal gaps in model performance and  demonstrate that even current frontier models struggle with faithful low-resource language summarization and evaluation in certain languages. For example, GPT-5.5's summaries of Nilo-Saharan documents are significantly less faithful than their English document counterparts. Native-speaker annotations validate the model judgments in our construction pipeline and judge faithfulness metrics. The proposed pipeline offers a general approach for repurposing existing sentence-level parallel resources for evaluation of document-level tasks.
\end{abstract}

\section{Introduction}

Due to the relative scarcity of text representing low-resource languages and the general cost of manual dataset construction, there are few long-form datasets available in such languages for evaluating long-form tasks such as summarization. The lack of evaluation resources beyond English and other high-resource languages has already been identified as a bottleneck in multilingual model development \citep{romanou2025include}. In the absence of these datasets, model capabilities and performance gaps in numerous languages remain unclear.

Typically, prior work relies on two primary families of methods for evaluating summaries, both of which pose challenges when applied to low-resource languages. First, \textit{reference-based metrics} compare model-generated summaries against human-written summaries (e.g., ROUGE \citep{lin2004rouge}, BLEURT \citep{sellam2020bleurt}).
While some low-resource languages have news paired with summaries (e.g., Moroccan Darija on Goud.ma \citep{issam2022goud}, Persian from BBC Persian \citep{hasan2021xlsum} and the languages of LR-Sum \citep{palenmichel2023lrsum}), coverage is uneven across languages and dialects, and collecting summaries can be prohibitively expensive at scale.

Second, \textit{reference-free metrics} verify a summary against the source document(s) (e.g., using NLI or LLM-as-a-Judge; \citet{laban2022summac,honovich2022true,min2023factscore}), alleviating the need for human-written references.
These metrics, however, presuppose that models or LLM judges sufficiently understand the language under evaluation. In either case, translation may be a natural approach to alleviating such resource scarcity, although similarly, \textit{translate-then-evaluate} approaches \citep{qiu2023mfact} then confound model performance with translation quality.

One method of addressing these concerns is through \textit{parallel long-form documents}: articles in a low-resource language paired with human-translated articles in English or another high-resource language. Assuming a capable summarizer in the high-resource language, comparing summaries produced from a pair of low resource/English articles
enables isolating the effect of input language on model performance. Similarly, this \textit{also} enables assessing the impact of input language on evaluation metrics and LLM judges.

These document-level parallel benchmarks do exist but are limited. NTREX-128 \citep{federmann2022ntrex}, for example, renders 123 news articles into 128 languages, including low-resource languages, but lacks representation of some dialects like Moroccan Arabic.
In contrast, \textit{sentence}-level parallel benchmarks, namely those intended for machine translation, are available for many languages and dialects that lack document-level evaluation data.
FLORES-200 \citep{nllb2022} alone translates the same 2,009 dev and devtest sentences into over 200 languages, MADAR \citep{bouamor2018madar} covers 25 Arabic city dialects, and community efforts such as SALT \citep{akera2022salt} provide tens of thousands of sentence pairs for Ugandan languages with little other digital text.

\textbf{Document-level Benchmarks from Parallel Sentences.}
In our paper, we leverage the availability of machine translation datasets to
address gaps in summarization evaluation in low-resource languages. We introduce an automated benchmark construction framework, \gls{approach}, which transforms parallel-sentence pairs into coherent documents for summarization benchmarks (Section~\ref{sec:pipeline}).
\gls{approach} first clusters translation sentence pairs using embedding models on the high-resource language side, then uses LLMs to order and filter sentences into coherent documents
(see Figure~\ref{fig:pipeline}).

Using \gls{approach}, we contribute a benchmark of 2{,}330 parallel articles across four Arabic dialects
and 1,341 in Lugbara, Acholi, Ateso, Persian, Dari and Tajik.\footnote{Lugbara (Central Sudanic), Acholi (Western Nilotic) and Ateso (Eastern Nilotic) are Nilo-Saharan languages of northwestern, northern and eastern Uganda; Persian, Dari and Tajik are the Iranian varieties of Iran, Afghanistan and Tajikistan, Tajik written in Cyrillic; the Arabic dialects are the regional groups of AraBench \citep{sajjad2020arabench} (Egyptian, Gulf, Levantine and Maghrebi).}
We validate a subset of the dialectal Arabic documents with judgments from native speakers; the other languages are validated automatically.

 Notably, the resulting documents are LLM-constructed but maintain the authenticity of human-written text in the original translation benchmark (Section~\ref{sec:validation-sec} and Appendix~\ref{app:nonllm}).
We utilize the resulting benchmark to investigate two research questions.

\textbf{RQ1) How faithfully do multilingual LLMs summarize documents in low-resource languages?} \\ We \textbf{assess cross-lingual summarization faithfulness}, using the constructed documents to evaluate whether LLMs summarize low-resource documents into English as well as they do English documents (Section~\ref{sec:rq3}). These experiments reveal performance gaps between models. In many cases large frontier models outperform specialized open-source models. Exceptions include Ateso and Lugbara. While scaling reasoning compute improves frontier model performance, a \textit{9 billion parameter model} fine-tuned on Lugbara slightly outperforms it.

\textbf{RQ2) Does LLM-as-a-Judge Performance Generalize to Low-Resource Languages?} \\
To measure summarization faithfulness, we previously evaluated summaries generated from low-resource source articles against parallel English source articles. We also \textbf{assess cross-lingual, LLM-based faithfulness evaluation}. Here, we evaluate whether an LLM can accurately verify claims against source documents in low-resource languages (Section~\ref{sec:rq2}).
In contrast to reliable performance on English source documents, we find that an accurate LLM judge can fail to correctly verify claims against low-resource source documents. These results highlight the importance of having access to parallel high-resource documents in low-resource settings, and future work on improving multilingual LLM-as-a-Judge performance in low-resource settings.

In summary, we propose \gls{approach}, a method for automatically constructing parallel documents from existing sentence-level machine translation datasets. We first use \gls{approach} to construct a cross-lingual summarization benchmark containing ten low-resource source languages (Section~\ref{sec:pipeline}). Evaluation on the benchmark reveals large performance gaps in both LLM summarization (Section~\ref{sec:rq3}) and judging capabilities (Section~\ref{sec:rq2}) across several languages.

\section{Related Work}
\label{sec:relatedwork}
\paragraph{Multilingual and low-resource summarization.}
Multilingual summarization datasets such as XL-Sum \citep{hasan2021xlsum}, CrossSum \citep{bhattacharjee2023crosssum}, MassiveSumm \citep{varab2021massivesumm}, and WikiLingua \citep{ladhak2020wikilingua} pair articles with summaries in many languages,
but they are harvested from web news and how-to guides, so they cover languages with substantial online text and rarely separate dialects. Evaluation built on FLORES-200 reaches far more languages, with short inputs and closed answers: Belebele asks multiple-choice questions about its passages in 122 language variants \citep{bandarkar2024belebele} and SIB-200 classifies its sentences by topic \citep{adelani2024sib200}, and suites such as AfroBench collect existing tasks for African languages \citep{ojo2025afrobench}. For Arabic, EASC \citep{elhaj2010easc} and KALIMAT \citep{elhaj2013kalimat} provide human-written and system-generated summaries of Modern Standard Arabic, respectively, dialect and culture benchmarks such as AraDiCE \citep{mousi2024aradice}, PALM \citep{alwajih2025palm} and DialectalArabicMMLU \citep{altakrori2026dialectalarabicmmlu} test knowledge through instruction--response pairs and multiple-choice questions, BALSAM \citep{almatham2025balsam} includes summarization among its tasks, and MArSum pairs 19{,}806 Moroccan Darija news articles with their headlines but has no English references to validate against \citep{gaanoun2022marsum}. None of these works pair long-form low-resource input with a parallel high-resource version to evaluate summarization faithfulness.

\paragraph{Faithfulness evaluation.}
Datasets and benchmarks such as FRANK \citep{pagnoni2021frank}, SummEval \citep{fabbri2021summeval}, TRUE \citep{honovich2022true}, AggreFact \citep{tang2023aggreFact}, and SummaC \citep{laban2022summac} assess whether generated summaries remain grounded in the source article.
FActScore improves this approach for longer text generations by decomposing outputs into smaller atomic claims and verifying the support for each claim \citep{min2023factscore}.
ReIFE \citep{liu2025reife} evaluates the LLM evaluators themselves across many base models and evaluation protocols, and finds that the base model matters more than the protocol.
Our judge follows both: a FActScore-inspired decompose-and-verify protocol run by a frontier model. JudgeBench \citep{tan2025judgebench} likewise builds a test of LLM judges from existing datasets with verifiable answers; ours checks the judge against the parallel English version.  Nevertheless, these resources are almost all English: SEAHORSE rates summaries in six languages, none of them among our ten \citep{clark2023seahorse}, and AraHalluEval evaluates Arabic without separating its dialects \citep{alansari2025arahallueval}.

\paragraph{Article-level data from sentence-level corpora.}

Other research has likewise attempted to generate article-level parallel data from sentence-level machine translation corpora. ParaDocs extracts article information from datasets like ParaCrawl, Europarl, and News Commentary to place the sentence pairs back into the article context from which they came \citep{wicks2023paradocs}; other studies use webpage or article metadata in the same way \citep{alghussin2023paracrawl,pal2024document}, and DocHPLT \citep{obrien2025dochplt} and DocBlocks \citep{ramos2025docblocks} maintain or create article-level parallel resources for long-context machine translation. These methods recover or preserve existing article structure. Earlier document-level MT work built ``fake articles'' from unrelated parallel sentences \citep{junczysdowmunt2019microsoft,berard2020naver}, but as training data, with no aim of coherence. For many low-resource languages and dialects no article-structured resource with references exists: the available data are collections of independent sentence pairs from phrasebooks, conversations or MT test sets, with no article structure to recover, and our pipeline composes new coherent articles from them while preserving authenticity to the source sentences.

\section{Automatic benchmark creation from parallel sentences}
\label{sec:pipeline}

We first introduce our framework for automatically constructing summarization data from sentence-level translation benchmarks, enabling further evaluation of LLMs' multilingual summarization and evaluation capabilities. We then use the scalable framework to construct a benchmark for cross-lingual summarization evaluation.

\begin{figure*}[t]
\centering
\resizebox{\textwidth}{!}{%
\input{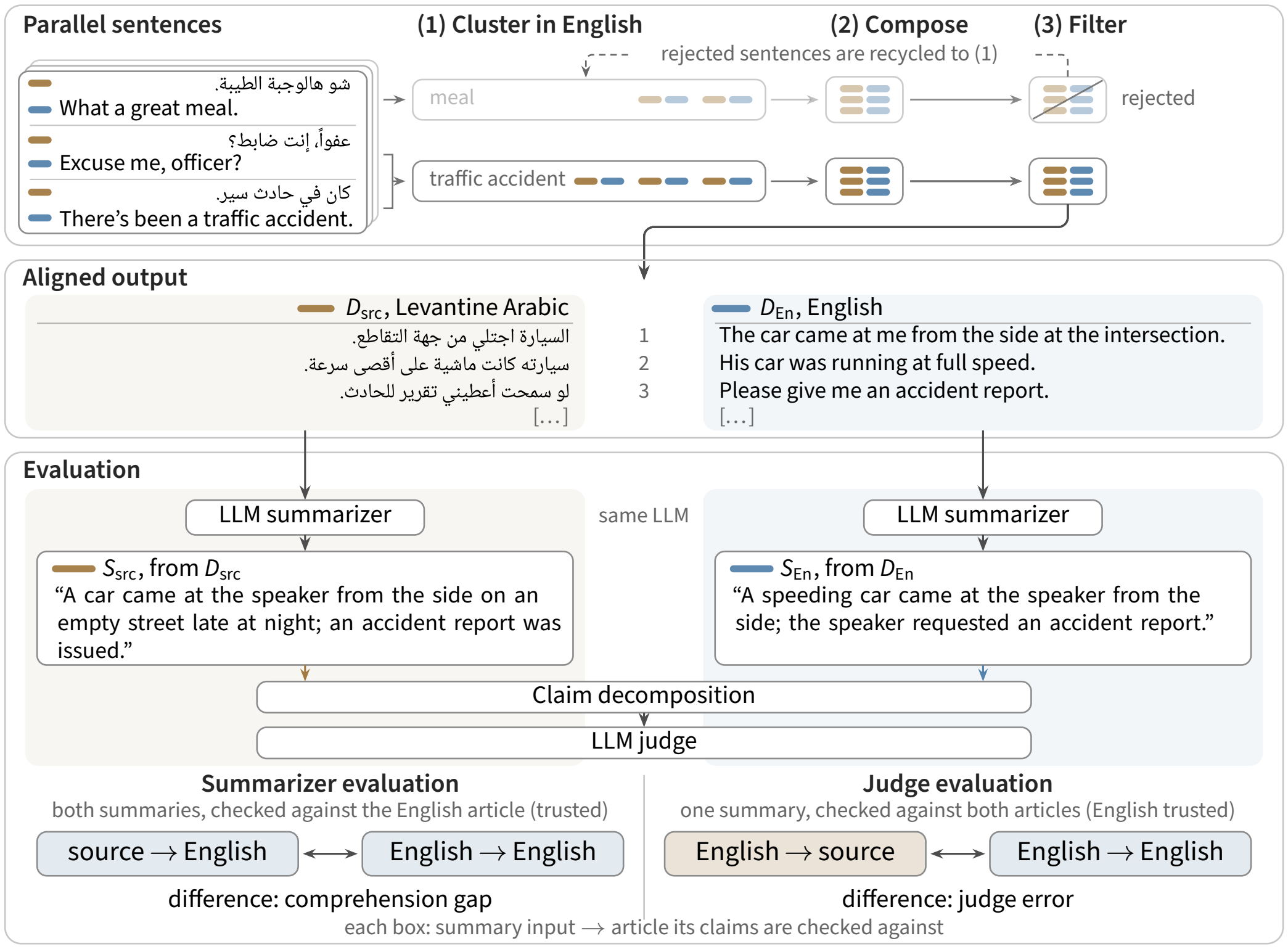}%
}
\caption{
\gls{approach} constructs articles from independent sentence pairs. Prompted English embeddings are clustered (1). An LLM selects and orders original pairs into articles (2). A separate LLM accepts coherent articles (3). Rejected candidates are recycled to step 1 and accepted documents form the aligned output. During evaluation, the same model summarizes each version into English. Following \citet{min2023factscore}, each summary is decomposed into claims, and an LLM judge checks the faithfulness of each claim. We first evaluate the faithfulness of generated summaries against the English document. We then evaluate cross-lingual LLM-as-a-Judge capabilities by comparing judged faithfulness scores from the English document to those computed against the low-resource reference document.
}
\label{fig:pipeline}
\end{figure*}

\subsection{\gls{approach}}

\paragraph{Problem Setting. }
We assume access to a sentence-level parallel dataset $D_{\mathrm{sentence}}=\{(a_i,b_i)\}_i$, where $a_i$ is a sentence in the low-resource language and $b_i$ is its gold human translation into a high-resource language. Our goal is to build  pairs of coherent multi-sentence articles from these sentences (Figure~\ref{fig:pipeline}). Each output article is a pair $(D_{\mathrm{src}},D_{\mathrm{En}})$ with $D_{\mathrm{src}}=(a_{i_1},\dots,a_{i_n})$ and $D_{\mathrm{En}}=(b_{i_1},\dots,b_{i_n})$ constructed from ordered sentence pairs.
The pipeline has three steps, numbered in Figure~\ref{fig:pipeline}.
We first cluster related pairs in the high-resource language (\textit{step 1}). We then compose candidate articles from each cluster (\textit{step 2}), filtering with a separate LLM (\textit{step 3}). Then we recycle the sentences of rejected candidates into further rounds (the loop back to step 1).
\paragraph{Step 1) Clustering.}

We first cluster unrelated pairs of sentences into groups of topically related sentences. We cluster on the high-resource side for better accuracy, producing clusters of 150 sentence pairs on average that could all refer to the same event.
 We embed texts with Qwen3-Embedding-8B \citep{zhang2025qwen3embedding}, prompted to capture whether or not the sentences could plausibly co-occur in the same document.
 Prompts
 and additional details are included in Appendices~\ref{app:prompts} and~\ref{app:models}.

\paragraph{Step 2) Composition.} A chosen frontier LLM (GPT-5.5; \citealp{openai2026gpt55}) takes each individual large cluster and is instructed to select, order, and divide the sentences into candidate articles. The composer prompt is reproduced in Appendix~\ref{app:prompts} and its settings in Appendix~\ref{app:models}.
Sentence pairs
are selected verbatim;
the one edit it may make is proper-noun unification, when two otherwise compatible sentences refer to different names, since independently written lines that fit one
event
can name different people. The principle is that the composer never adds content from its own knowledge.

\paragraph{Step 3) Filtering.}
\label{sec:gating}

Not every candidate produced by the generation model is coherent. We therefore apply a separate frontier LLM-as-a-Judge filter for coherence (Claude Sonnet 4.6; \citealp{anthropic2026sonnet46}).
A different model is given each constructed article and instructed to rate its overall coherence on a scale from 1 to 5. We define coherence as whether the sentences flow naturally into one article. The
prompt for filtering is reproduced in Appendix~\ref{app:prompts}. We keep articles that receive a coherence score of at least 4. Any article scoring below 4 is rejected, broken back down into its original sentence pairs, and returned to the candidate pool. On a sample of 32 accepted and rejected articles, the filter agrees with native speakers' average rating at Cohen's $\kappa=0.62$ (Appendix~\ref{app:human}).

Finally, we take the rejected pool of sentences and repeat all steps from scratch.
Rejected sentences can be reclustered into different groups, allowing them to form new coherent articles. Each complete iteration of this process is considered one round. The full process ends either after a chosen maximum number of rounds or when the passing rate falls below a selected threshold. The maximum number of rounds and the stopping threshold used for each build are listed in Appendix~\ref{app:models}. The later rounds produce most of the articles (Table~\ref{tab:recycling} in Appendix~\ref{app:transfer}), and their articles pass through the same coherence filter as the first round's, so they meet the same quality bar.

\subsection{A Ten-Language Summarization Benchmark}

\gls{approach} enables constructing pairs of synthetic documents for any low-resource language with a high-resource sentence-level machine translation dataset. Notably, these datasets are readily available for many languages and dialects. We utilize the method to construct a benchmark of ten languages that is later used for multilingual summarization evaluation.

\label{sec:transfer} We ran our pipeline across three language families with existing parallel corpora (Table~\ref{tab:arabic-instantiation}):

 \begin{itemize}[leftmargin=10pt]
     \item \textit{Afro-Asiatic}: We use four Arabic dialect groups from 105{,}590 dialect-English pairs \citep{bouamor2018madar,sajjad2020arabench}. This produced 2{,}330 articles (301 Egyptian, 914 Gulf, 435 Levantine and 680 Maghrebi).
     \item  \textit{Nilo-Saharan}: We use Lugbara, a Central Sudanic language of Uganda, from 24{,}946 Lugbara-English pairs of SALT \citep{akera2022salt}, giving 240 articles whose multi-way translations give the same articles in Acholi and Ateso, two Nilotic languages of the same family.
     \item  \textit{Indo-Iranian branch of the Indo-European family}: We use Persian, Dari and Tajik, from the 2{,}009 dev and devtest sentences of FLORES-200 \citep{nllb2022}, translated from English, giving 232, 230 and 159 articles.
 \end{itemize}
 In total, the resulting dataset has 3{,}671 articles, of which 2{,}330 are in the Arabic dialects and 1,341 are in the other two families.
These languages span multiple resource classes.  All three Nilo-Saharan languages are in the lowest resource class of \citet{joshi2020state}, whereas Persian is in class 4 and Egyptian Arabic in class 3.

SALT's larger pool also allows longer articles (15 sentences on average); on the Arabic articles, length has little effect on faithfulness (Appendix~\ref{app:length}). The FLORES varieties also serve as a control because their sentences come from known pages in fragments of two to five sentences (Appendix~\ref{app:flores}). We validate the resulting dataset via a native-speaker study. We include additional automatic metrics in Appendix~\ref{app:ablation}.

\label{sec:validation-sec}

\paragraph{Human Benchmark Validation.}
\label{sec:validation}

We conducted a native-speaker study of synthetic documents in four Arabic dialects with assistance from crowd workers.\footnote{The study was reviewed and approved by the Columbia University IRB. Annotators gave informed consent and were paid 17 USD per hour.} Seven native Arabic dialect speakers
participated
(two Egyptian, two Levantine, two Maghrebi, and one Gulf). Annotators evaluated the quality of the accepted synthetic articles against
natural news, shuffled variants of the accepted articles, and rejected articles from the earlier coherence gate in step 3 (Figure~\ref{fig:pipeline}; Section~\ref{sec:gating}).

Every annotator completed three tasks on articles in their own dialect (full task descriptions are shown in Appendix~\ref{app:human}).
Annotators (1) evaluated the \gls{approach} composition step by identifying a random sentence from sentences in the same constructed document,   (2) completed a quality checklist to evaluate overall article quality, and (3) provided coherence and human-likeness ratings for each type of document.

\begin{table}[t]
\centering
\small
\renewcommand{\arraystretch}{0.86}
\begin{tabular}{lcccc}
\toprule
& \textbf{Natural news} & \textbf{Accepted} & \textbf{Rejected} & \textbf{Shuffled} \\
\midrule
Coherence (1-4) & 3.93 & 3.07 & 2.11 & 2.04 \\
Human-written (1-4) & 3.75 & 2.57 & 2.25 & 2.18 \\
\bottomrule
\end{tabular}
\caption{Native-speaker ratings by hidden condition (7 annotators): natural news, articles the gate accepted, articles the gate rejected (not in the benchmark), and accepted articles with their sentence order shuffled.}
\label{tab:cp2}
\end{table}
\paragraph{Human Validation Results.}  Every annotator identified the intruder in all eight odd-one-out items from task 1. On the task 3 ratings (Table~\ref{tab:cp2}), natural news scores highest, accepted articles next, and the two other controls the lowest as expected. Annotators most often faulted rejected articles as ``shared setting only'' (independent lines from one domain rather than one continuous scene), exactly what the gate was built to reject. The annotators marked shuffled ones for broken order. Accepted and shuffled articles contain the same sentences, so the difference between their ratings is the order the composer imposed. Accepted articles still score below natural news, which we partially attribute to their source: MADAR sentences are conversational phrasebook lines, so an accepted article reads like a dialogue rather than a news article. By contrast, the coherence gate rates the Nilo-Saharan articles, built from written prose, as much more coherent, likely owing to their source register (Appendix~\ref{app:qualitative}).

\section{RQ1) How Well Do LLMs Summarize Low-Resource Languages?}
\label{sec:rq3}
\label{sec:results}

In this section, we evaluate the quality of summaries generated by LLMs from low-resource source documents. We focus on evaluating summary \textit{faithfulness} to the source document.

\paragraph{Summaries and claims.}
\label{sec:method}
Each system generates an English summary
for every article under two input conditions: the input article in the low-resource language, or the English article formed from the gold translations. We limit summary generation to 30 words to ensure comparable, one-sentence summaries across systems.
Each summary is broken into atomic claims by the LLM judge, and each claim is checked against the English article as reference. We write $F_{r,g}$ for the faithfulness of the summary written from article $r$ when its claims are checked against article $g$ (Figure~\ref{fig:pipeline}).

\paragraph{Judging Faithfulness Against English Source Documents.}
Each summary claim receives one of four labels: \textit{supported}, \textit{partially supported}, \textit{not supported}, or \textit{contradicted} by the LLM-as-a-Judge. Faithfulness is the fraction of supported claims with half credit for partial support: \footnote{Half credit is empirically validated; scoring the annotators' ratings of the claims the judge marks partially supported with the same weights gives them 0.63 credit on average (Appendix~\ref{app:human}).}

\begin{equation}
F_{r,g}=\frac{N^{r,g}_{\mathrm{sup}}+\tfrac12\,N^{r,g}_{\mathrm{par}}}{N^{r}_{\mathrm{claims}}},
\label{eq:faith}
\end{equation}
where $N^{r,g}_{\mathrm{sup}}$ and $N^{r,g}_{\mathrm{par}}$ count the claims of the summary generated from input $r$ that reference $g$ supports fully or partially, and $N^{r}_{\mathrm{claims}}$ counts all claims of that summary. Figure~\ref{fig:claims} in Appendix~\ref{app:qualitative} shows the claims of one article's two summaries, each checked for consistency with each reference article.
We freeze one decomposition per summary and judge identical claims under both languages. All judging is performed by Claude Sonnet 4.6. Here, we build on existing work that has demonstrated the effectiveness of LLM-as-a-Judge for evaluating English summaries of English documents \citep{song2024finesure}. We also validate the judge in this specific setting with human annotations.

As part of the native-speaker study, human annotators also classified whether claims extracted from generated summaries were faithful or unfaithful to the source document. Claude Sonnet 4.6 and annotators agreed on 70\% of claims ($\kappa=0.40$), close to the inter-annotator agreement (73\%, $\kappa=0.45$). This suggests that much of the remaining disagreement is noise rather than a systematic judge error.
 We additionally evaluate faithfulness with AlignScore \citep{zha2023alignscore} and MiniCheck \citep{tang2024minicheck}, which correlate with the LLM-as-a-Judge on English documents (Appendices~\ref{app:judge} and~\ref{app:anchors}). Together, these results provide evidence that LLM-as-a-Judge scores of faithfulness are reliable when comparing English summaries with English source documents.
 We therefore use this judge, along with the English source document, to evaluate the faithfulness of summaries generated from low-resource language documents.

\subsection{Experiments}

We evaluate the faithfulness of summaries generated from both the low-resource language document and its corresponding English translation against the English article (Figure~\ref{fig:pipeline}). We use \textit{faithfulness gap} to refer to the difference in scores between summaries generated from the high-resource and low-resource documents, $F_{\mathrm{En},\mathrm{En}}$ and $F_{\mathrm{src},\mathrm{En}}$.

\paragraph{Models.}
We evaluate six systems on faithful summarization: GPT-5.5, gpt-oss-20B \citep{openai2025gptoss}, Qwen3-30B-A3B and Qwen3-8B \citep{yang2025qwen3} as open general models, Fanar-1-9B \citep{fanar2025} as an Arabic-specialized model, and Aya-23-8B \citep{aryabumi2024aya23} as a multilingual model. On the Nilo-Saharan languages, we use the specialists Sunflower-Qwen3.5-9B \citep{sunbird2026sunflower9b} and Sunflower-32B \citep{akera2026sunflower}, and for the Persian varieties, the regional specialist Dorna2-8B \citep{partai2024dorna2}.
On Lugbara, we also run GPT-5.6-Sol \citep{openai2026gpt56} at each of its five reasoning efforts.

\begin{figure}[t]
\centering
\begingroup
\begin{tikzpicture}[x=1cm,y=1cm,font=\sffamily\fontsize{8}{9}\selectfont,every node/.style={inner sep=0pt}]
\definecolor{gapGPT}{HTML}{202B3C}
\definecolor{gapQwen30}{HTML}{0072B2}
\definecolor{gapOSS}{HTML}{D55E00}
\definecolor{gapQwen8}{HTML}{7757A8}
\definecolor{gapAya}{HTML}{BE507E}
\definecolor{gapSpecialist}{HTML}{008875}
\definecolor{gapGrid}{HTML}{E4E8ED}
\definecolor{gapText}{HTML}{263241}
\definecolor{gapMuted}{HTML}{647181}
\path[use as bounding box] (0.000,-0.450) rectangle (13.278,5.637);
\newcommand{\gapmark}[4]{%
  \begin{scope}[shift={(#1,#2)}]
  \ifcase#3\relax
    \filldraw[fill=#4,draw=white,line width=.3pt] (0.000,0.000) circle (.058);
  \or
    \filldraw[fill=#4,draw=white,line width=.3pt] (0.000,0.071) -- (-0.067,-0.045) -- (0.067,-0.045) -- cycle;
  \or
    \filldraw[fill=#4,draw=white,line width=.3pt] (-0.054,-0.054) rectangle (0.054,0.054);
  \or
    \filldraw[fill=#4,draw=white,line width=.3pt] (0.000,0.072) -- (0.066,0.000) -- (0.000,-0.072) -- (-0.066,0.000) -- cycle;
  \or
    \filldraw[fill=#4,draw=white,line width=.3pt] (0.000,-0.071) -- (-0.067,0.045) -- (0.067,0.045) -- cycle;
  \or
    \filldraw[fill=#4,draw=white,line width=.3pt] (0.000,0.082) -- (0.019,0.026) -- (0.078,0.025) -- (0.031,-0.010) -- (0.048,-0.066) -- (0.000,-0.032) -- (-0.048,-0.066) -- (-0.031,-0.010) -- (-0.078,0.025) -- (-0.019,0.026) -- cycle;
  \or
    \filldraw[fill=white,draw=#4,line width=.6pt] (0.000,0.082) -- (0.019,0.026) -- (0.078,0.025) -- (0.031,-0.010) -- (0.048,-0.066) -- (0.000,-0.032) -- (-0.048,-0.066) -- (-0.031,-0.010) -- (-0.078,0.025) -- (-0.019,0.026) -- cycle;
  \fi
  \end{scope}%
}
\gapmark{2.100}{5.427}{0}{gapGPT}
\node[anchor=west,text=gapText] at (2.247,5.427) {GPT-5.5};
\gapmark{4.906}{5.427}{1}{gapQwen30}
\node[anchor=west,text=gapText] at (5.053,5.427) {Qwen3-30B-A3B};
\gapmark{7.712}{5.427}{2}{gapOSS}
\node[anchor=west,text=gapText] at (7.859,5.427) {gpt-oss-20B};
\gapmark{10.518}{5.427}{3}{gapQwen8}
\node[anchor=west,text=gapText] at (10.665,5.427) {Qwen3-8B};
\gapmark{2.100}{5.047}{4}{gapAya}
\node[anchor=west,text=gapText] at (2.247,5.047) {Aya-23-8B};
\gapmark{4.906}{5.047}{5}{gapSpecialist}
\node[anchor=west,text=gapText] at (5.053,5.047) {Specialist};
\gapmark{7.712}{5.047}{6}{gapSpecialist}
\node[anchor=west,text=gapText] at (7.859,5.047) {Sunflower-32B};
\fill[black!2] (0.020,4.255) rectangle (13.094,4.660);
\node[anchor=east,text=gapText] at (1.880,4.458) {Persian};
\node[anchor=east,text=gapText] at (1.880,4.052) {Dari};
\fill[black!2] (0.020,3.444) rectangle (13.094,3.849);
\node[anchor=east,text=gapText] at (1.880,3.647) {Tajik};
\node[anchor=east,text=gapText] at (1.880,3.241) {Egyptian};
\fill[black!2] (0.020,2.632) rectangle (13.094,3.038);
\node[anchor=east,text=gapText] at (1.880,2.835) {Gulf};
\node[anchor=east,text=gapText] at (1.880,2.430) {Levantine};
\fill[black!2] (0.020,1.821) rectangle (13.094,2.227);
\node[anchor=east,text=gapText] at (1.880,2.024) {Maghrebi};
\node[anchor=east,text=gapText] at (1.880,1.619) {Lugbara};
\fill[black!2] (0.020,1.010) rectangle (13.094,1.416);
\node[anchor=east,text=gapText] at (1.880,1.213) {Acholi};
\node[anchor=east,text=gapText] at (1.880,0.807) {Ateso};
\node[rotate=90,text=gapMuted,font=\sffamily\fontsize{6.5}{7}\selectfont] at (0.200,4.052) {Persian};
\node[rotate=90,text=gapMuted,font=\sffamily\fontsize{6.5}{7}\selectfont] at (0.200,2.632) {Arabic};
\node[rotate=90,text=gapMuted,font=\sffamily\fontsize{6.5}{7}\selectfont] at (0.200,1.213) {Nilo-Saharan};
\draw[gapGrid,line width=.35pt] (2.100,0.585) -- (2.100,4.672);
\node[anchor=north,text=gapMuted] at (2.100,0.355) {0.0};
\draw[gapGrid,line width=.35pt] (3.457,0.585) -- (3.457,4.672);
\node[anchor=north,text=gapMuted] at (3.457,0.355) {0.1};
\draw[gapGrid,line width=.35pt] (4.814,0.585) -- (4.814,4.672);
\node[anchor=north,text=gapMuted] at (4.814,0.355) {0.2};
\draw[gapGrid,line width=.35pt] (6.171,0.585) -- (6.171,4.672);
\node[anchor=north,text=gapMuted] at (6.171,0.355) {0.3};
\draw[gapGrid,line width=.35pt] (7.528,0.585) -- (7.528,4.672);
\node[anchor=north,text=gapMuted] at (7.528,0.355) {0.4};
\draw[gapGrid,line width=.35pt] (8.885,0.585) -- (8.885,4.672);
\node[anchor=north,text=gapMuted] at (8.885,0.355) {0.5};
\draw[gapGrid,line width=.35pt] (10.242,0.585) -- (10.242,4.672);
\node[anchor=north,text=gapMuted] at (10.242,0.355) {0.6};
\draw[gapGrid,line width=.35pt] (11.599,0.585) -- (11.599,4.672);
\node[anchor=north,text=gapMuted] at (11.599,0.355) {0.7};
\draw[gapGrid,line width=.35pt] (12.956,0.585) -- (12.956,4.672);
\node[anchor=north,text=gapMuted] at (12.956,0.355) {0.8};
\draw[gapMuted,line width=.65pt] (2.100,0.585) -- (2.100,4.672);
\draw[gapGrid,line width=.45pt] (2.100,0.585) -- (12.956,0.585);
\draw[gapMuted,line width=.6pt] (0.420,3.444) -- (13.094,3.444);
\draw[gapMuted,line width=.6pt] (0.420,1.821) -- (13.094,1.821);
\gapmark{2.507}{4.458}{0}{gapGPT} %
\gapmark{3.457}{4.458}{1}{gapQwen30} %
\gapmark{4.000}{4.458}{2}{gapOSS} %
\gapmark{3.864}{4.458}{3}{gapQwen8} %
\gapmark{2.779}{4.458}{4}{gapAya} %
\gapmark{3.186}{4.458}{5}{gapSpecialist} %
\gapmark{2.643}{4.052}{0}{gapGPT} %
\gapmark{3.728}{4.052}{1}{gapQwen30} %
\gapmark{4.543}{4.052}{2}{gapOSS} %
\gapmark{4.407}{4.052}{3}{gapQwen8} %
\gapmark{3.321}{4.052}{4}{gapAya} %
\gapmark{4.271}{4.052}{5}{gapSpecialist} %
\gapmark{2.507}{3.647}{0}{gapGPT} %
\gapmark{5.085}{3.647}{1}{gapQwen30} %
\gapmark{6.307}{3.647}{2}{gapOSS} %
\gapmark{5.900}{3.647}{3}{gapQwen8} %
\gapmark{11.328}{3.647}{4}{gapAya} %
\gapmark{5.628}{3.647}{5}{gapSpecialist} %
\gapmark{2.914}{3.241}{0}{gapGPT} %
\gapmark{4.407}{3.241}{1}{gapQwen30} %
\gapmark{3.864}{3.241}{2}{gapOSS} %
\gapmark{4.814}{3.241}{3}{gapQwen8} %
\gapmark{3.457}{3.241}{4}{gapAya} %
\gapmark{3.728}{3.241}{5}{gapSpecialist} %
\gapmark{3.050}{2.835}{0}{gapGPT} %
\gapmark{4.814}{2.835}{1}{gapQwen30} %
\gapmark{4.271}{2.835}{2}{gapOSS} %
\gapmark{5.357}{2.835}{3}{gapQwen8} %
\gapmark{4.407}{2.777}{4}{gapAya} %
\gapmark{4.407}{2.894}{5}{gapSpecialist} %
\gapmark{3.050}{2.430}{0}{gapGPT} %
\gapmark{5.085}{2.430}{1}{gapQwen30} %
\gapmark{4.543}{2.430}{2}{gapOSS} %
\gapmark{5.492}{2.430}{3}{gapQwen8} %
\gapmark{4.271}{2.430}{4}{gapAya} %
\gapmark{4.678}{2.430}{5}{gapSpecialist} %
\gapmark{3.321}{2.024}{0}{gapGPT} %
\gapmark{6.307}{2.024}{1}{gapQwen30} %
\gapmark{4.950}{2.024}{2}{gapOSS} %
\gapmark{6.442}{2.024}{3}{gapQwen8} %
\gapmark{5.357}{2.024}{4}{gapAya} %
\gapmark{5.492}{2.024}{5}{gapSpecialist} %
\gapmark{7.935}{1.619}{0}{gapGPT} %
\gapmark{11.192}{1.619}{1}{gapQwen30} %
\gapmark{8.071}{1.619}{2}{gapOSS} %
\gapmark{9.835}{1.619}{3}{gapQwen8} %
\gapmark{12.142}{1.619}{4}{gapAya} %
\gapmark{6.714}{1.619}{5}{gapSpecialist} %
\gapmark{5.492}{1.619}{6}{gapSpecialist} %
\gapmark{4.271}{1.213}{0}{gapGPT} %
\gapmark{11.735}{1.213}{1}{gapQwen30} %
\gapmark{9.971}{1.213}{2}{gapOSS} %
\gapmark{9.292}{1.213}{3}{gapQwen8} %
\gapmark{12.549}{1.213}{4}{gapAya} %
\gapmark{4.678}{1.271}{5}{gapSpecialist} %
\gapmark{4.678}{1.154}{6}{gapSpecialist} %
\gapmark{5.900}{0.807}{0}{gapGPT} %
\gapmark{11.870}{0.807}{1}{gapQwen30} %
\gapmark{9.292}{0.807}{2}{gapOSS} %
\gapmark{9.835}{0.807}{3}{gapQwen8} %
\gapmark{12.820}{0.807}{4}{gapAya} %
\gapmark{4.543}{0.807}{5}{gapSpecialist} %
\gapmark{4.950}{0.807}{6}{gapSpecialist} %
\node[anchor=north,text=gapText] at (7.528,-0.045) {Faithfulness gap (English-input faithfulness $-$ source-input faithfulness)};
\end{tikzpicture}
\endgroup
\caption{Each summarization system's faithfulness with English input minus its faithfulness with source-language input. Both are judged against the English article (0 means equally faithful), with rules separating the three language families. The specialist is Dorna2-8B for the Persian varieties, Fanar-1-9B for the Arabic dialects and Sunflower-Qwen3.5-9B for the Nilo-Saharan languages, which also show Sunflower-32B (hollow stars). Table~\ref{tab:2x2} provides every value.
}
\label{fig:summ}
\end{figure}
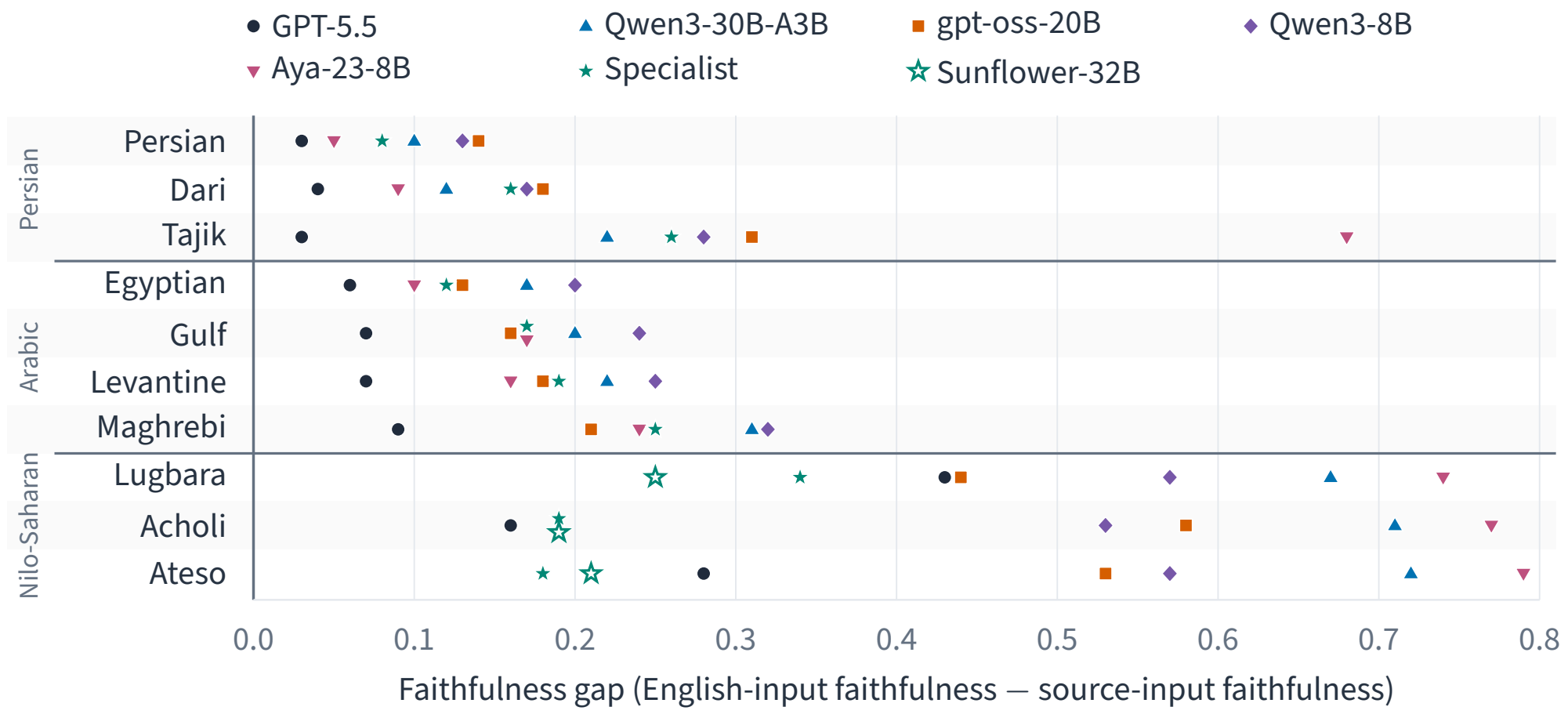

\textbf{Results.} Table~\ref{tab:2x2} in Appendix~\ref{app:dialect} contains faithfulness scores for each summarization system. On all four Arabic dialects and all three Persian varieties evaluated,
GPT-5.5 outperforms the corresponding specialized language models. The notable exception is the Nilo-Saharan languages, where the specialist language models are most performant in two of three languages. The Sunflower specialist models are the most faithful systems on Lugbara and Ateso, outperforming all other models. Notably, on Lugbara even \textit{Sunflower-9B} slightly outperforms GPT-5.5 (0.50 to 0.47).

The strongest frontier model loses much of its faithfulness when the input article is in a Nilo-Saharan language: GPT-5.5 scores 0.90 with English input, but 0.47 with Lugbara, 0.74 with Acholi and 0.62 with Ateso input. Summarization from very low-resource languages therefore remains unsolved even for frontier models, and without a parallel English article this loss would be hard to measure, since the judge cannot reliably verify claims against the Lugbara or Ateso article.

Figure~\ref{fig:summ} plots the difference in faithfulness scores when summarizing from English versus the low-resource language. Every system is on average less faithful when summarizing from the low-resource document. The difference is minimal for GPT-5.5 on the Persian varieties and Arabic dialects (at most 0.09), but larger for the four general open-weight models (0.21 on average), and these losses track how well each system translates the source sentences (Appendix~\ref{app:translation}). The difference for GPT-5.5 is largest on the Nilo-Saharan languages, where faithfulness roughly halves when generating summaries from Lugbara documents ($0.74$ and $0.62$ in Acholi and Ateso). In reasoning intensity experiments with GPT-5.6-Sol, faithfulness with Lugbara input rises with increasing reasoning effort, from $0.37$ without reasoning to $0.49$ with reasoning set to xhigh (Figure~\ref{fig:effort} in Appendix~\ref{app:transfer}).

Table~\ref{tab:errortags} in Appendix~\ref{app:errors} shows the claim error types annotated by the LLM-as-a-Judge. On the Arabic dialects, summaries including incorrect names and wrong numbers become several times as frequent when the input article is in the dialect. In Lugbara, models are more likely to include unsupported claims, twelve times the English-input rate. Additionally, the summaries themselves change style, and overgeneralize. Half of gpt-oss-20B's Lugbara-input summaries and a quarter of Qwen3-8B's yield no claim.  Overall, the faithfulness of summaries generated from documents in low-resource languages is lower than those generated from English document counterparts, with error types dependent on source language. Next, we evaluate the effect of source language on LLM-as-a-Judge performance.

\section{RQ2) Does LLM-as-a-Judge Performance Generalize to Low-Resource Languages?}
\label{sec:rq2}

In this section, we study whether LLM-as-a-Judge capabilities are maintained when checking claims against reference documents in low-resource languages.

In the previous section, we found that Claude Sonnet 4.6 was a reliable judge of faithfulness, aligning with previous work on LLM judges for high-resource summarization tasks \citep{song2024finesure}. To evaluate whether this capability generalizes to other languages, we compare performance when utilizing the low-resource text as the reference document.

\subsection{Experiments}
We use the summaries generated from English source documents in the summarization evaluation experiment. As before, we extract claims from each generated English summary. However, in contrast to the summary faithfulness evaluation in the previous section, we now use the Claude Sonnet 4.6 LLM-as-a-Judge to check claims against \textit{both} the low-resource and English versions of each source document (Figure~\ref{fig:pipeline}, bottom). We examine whether claim judgments change based on reference document language, treating the verdict against the English document as correct. We consider \textit{claim agreement} and the difference in average faithfulness scores.

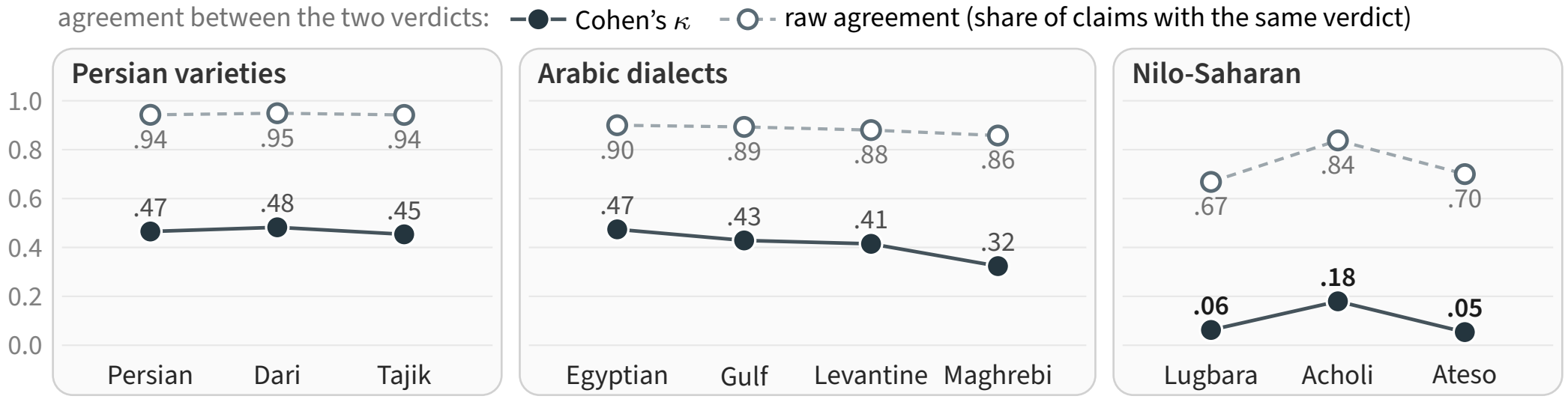
\begin{figure}[t]
\centering
\begin{tikzpicture}[x=1cm,y=1cm,font=\sffamily\scriptsize,every node/.style={inner sep=1pt},
  panel/.style={draw=black!18,line width=.5pt,rounded corners=5pt,fill=black!2},
  head/.style={font=\sffamily\bfseries\fontsize{7.6}{9}\selectfont,text=black!80},
  grid/.style={draw=black!9,line width=.4pt},
  kline/.style={draw=cbInk!85,line width=.9pt},
  aline/.style={draw=cbMuted!60,line width=.8pt,dash pattern=on 2.4pt off 1.8pt},
  kdot/.style={circle,fill=cbInk,draw=white,line width=.6pt,inner sep=0pt,minimum size=5.6pt},
  adot/.style={circle,fill=white,draw=cbMuted,line width=.9pt,inner sep=0pt,minimum size=4.6pt}]
\draw[panel] (0.55,-0.44) rectangle (4.51,2.62);
\node[head,anchor=north west] at (0.67,2.55) {Persian varieties};
\draw[panel] (4.67,-0.44) rectangle (9.75,2.62);
\node[head,anchor=north west] at (4.79,2.55) {Arabic dialects};
\draw[panel] (9.91,-0.44) rectangle (13.87,2.62);
\node[head,anchor=north west] at (10.03,2.55) {Nilo-Saharan};
\draw[grid] (0.63,0.000) -- (4.43,0.000);
\draw[grid] (4.75,0.000) -- (9.67,0.000);
\draw[grid] (9.99,0.000) -- (13.79,0.000);
\node[anchor=east,text=black!50] at (0.50,0.000) {0.0};
\draw[grid] (0.63,0.432) -- (4.43,0.432);
\draw[grid] (4.75,0.432) -- (9.67,0.432);
\draw[grid] (9.99,0.432) -- (13.79,0.432);
\node[anchor=east,text=black!50] at (0.50,0.432) {0.2};
\draw[grid] (0.63,0.864) -- (4.43,0.864);
\draw[grid] (4.75,0.864) -- (9.67,0.864);
\draw[grid] (9.99,0.864) -- (13.79,0.864);
\node[anchor=east,text=black!50] at (0.50,0.864) {0.4};
\draw[grid] (0.63,1.296) -- (4.43,1.296);
\draw[grid] (4.75,1.296) -- (9.67,1.296);
\draw[grid] (9.99,1.296) -- (13.79,1.296);
\node[anchor=east,text=black!50] at (0.50,1.296) {0.6};
\draw[grid] (0.63,1.728) -- (4.43,1.728);
\draw[grid] (4.75,1.728) -- (9.67,1.728);
\draw[grid] (9.99,1.728) -- (13.79,1.728);
\node[anchor=east,text=black!50] at (0.50,1.728) {0.8};
\draw[grid] (0.63,2.160) -- (4.43,2.160);
\draw[grid] (4.75,2.160) -- (9.67,2.160);
\draw[grid] (9.99,2.160) -- (13.79,2.160);
\node[anchor=east,text=black!50] at (0.50,2.160) {1.0};
\draw[aline] (1.410,2.036) -- (2.530,2.050) -- (3.650,2.035);
\draw[kline] (1.410,1.005) -- (2.530,1.042) -- (3.650,0.980);
\node[adot] at (1.410,2.036) {};
\node[anchor=north,text=black!55] at (1.410,1.946) {.94};
\node[kdot] at (1.410,1.005) {};
\node[anchor=south,text=black!70] at (1.410,1.095) {.47};
\node[anchor=north,text=black!85] at (1.410,-0.130) {Persian};
\node[adot] at (2.530,2.050) {};
\node[anchor=north,text=black!55] at (2.530,1.960) {.95};
\node[kdot] at (2.530,1.042) {};
\node[anchor=south,text=black!70] at (2.530,1.132) {.48};
\node[anchor=north,text=black!85] at (2.530,-0.130) {Dari};
\node[adot] at (3.650,2.035) {};
\node[anchor=north,text=black!55] at (3.650,1.945) {.94};
\node[kdot] at (3.650,0.980) {};
\node[anchor=south,text=black!70] at (3.650,1.070) {.45};
\node[anchor=north,text=black!85] at (3.650,-0.130) {Tajik};
\draw[aline] (5.530,1.944) -- (6.650,1.929) -- (7.770,1.901) -- (8.890,1.853);
\draw[kline] (5.530,1.024) -- (6.650,0.926) -- (7.770,0.896) -- (8.890,0.699);
\node[adot] at (5.530,1.944) {};
\node[anchor=north,text=black!55] at (5.530,1.854) {.90};
\node[kdot] at (5.530,1.024) {};
\node[anchor=south,text=black!70] at (5.530,1.114) {.47};
\node[anchor=north,text=black!85] at (5.530,-0.130) {Egyptian};
\node[adot] at (6.650,1.929) {};
\node[anchor=north,text=black!55] at (6.650,1.839) {.89};
\node[kdot] at (6.650,0.926) {};
\node[anchor=south,text=black!70] at (6.650,1.016) {.43};
\node[anchor=north,text=black!85] at (6.650,-0.130) {Gulf};
\node[adot] at (7.770,1.901) {};
\node[anchor=north,text=black!55] at (7.770,1.811) {.88};
\node[kdot] at (7.770,0.896) {};
\node[anchor=south,text=black!70] at (7.770,0.986) {.41};
\node[anchor=north,text=black!85] at (7.770,-0.130) {Levantine};
\node[adot] at (8.890,1.853) {};
\node[anchor=north,text=black!55] at (8.890,1.763) {.86};
\node[kdot] at (8.890,0.699) {};
\node[anchor=south,text=black!70] at (8.890,0.789) {.32};
\node[anchor=north,text=black!85] at (8.890,-0.130) {Maghrebi};
\draw[aline] (10.770,1.443) -- (11.890,1.810) -- (13.010,1.512);
\draw[kline] (10.770,0.135) -- (11.890,0.388) -- (13.010,0.116);
\node[adot] at (10.770,1.443) {};
\node[anchor=north,text=black!55] at (10.770,1.353) {.67};
\node[kdot] at (10.770,0.135) {};
\node[anchor=south,font=\sffamily\bfseries\scriptsize,text=black!90] at (10.770,0.225) {.06};
\node[anchor=north,text=black!85] at (10.770,-0.130) {Lugbara};
\node[adot] at (11.890,1.810) {};
\node[anchor=north,text=black!55] at (11.890,1.720) {.84};
\node[kdot] at (11.890,0.388) {};
\node[anchor=south,font=\sffamily\bfseries\scriptsize,text=black!90] at (11.890,0.478) {.18};
\node[anchor=north,text=black!85] at (11.890,-0.130) {Acholi};
\node[adot] at (13.010,1.512) {};
\node[anchor=north,text=black!55] at (13.010,1.422) {.70};
\node[kdot] at (13.010,0.116) {};
\node[anchor=south,font=\sffamily\bfseries\scriptsize,text=black!90] at (13.010,0.206) {.05};
\node[anchor=north,text=black!85] at (13.010,-0.130) {Ateso};
\node[anchor=west,text=black!55] (l0) at (0.55,2.88) {agreement between the two verdicts:};
\draw[kline] ([xshift=4pt]l0.east) -- ([xshift=18pt]l0.east);
\node[kdot] (l1) at ([xshift=11pt]l0.east) {};
\node[anchor=west] (l2) at ([xshift=19pt]l0.east) {Cohen's $\kappa$};
\draw[aline] ([xshift=6pt]l2.east) -- ([xshift=20pt]l2.east);
\node[adot] (l3) at ([xshift=13pt]l2.east) {};
\node[anchor=west] at ([xshift=21pt]l2.east) {raw agreement (share of claims with the same verdict)};
\end{tikzpicture}%
\caption{Comparison of LLM-as-a-Judge (Claude Sonnet 4.6) verdicts on summary claims checked against the source-language article versus the English article. $\kappa$ is 0.32 to 0.48 for the Persian varieties and Arabic dialects and at most 0.18 for the Nilo-Saharan languages.}
\label{fig:judge-agree}
\end{figure}

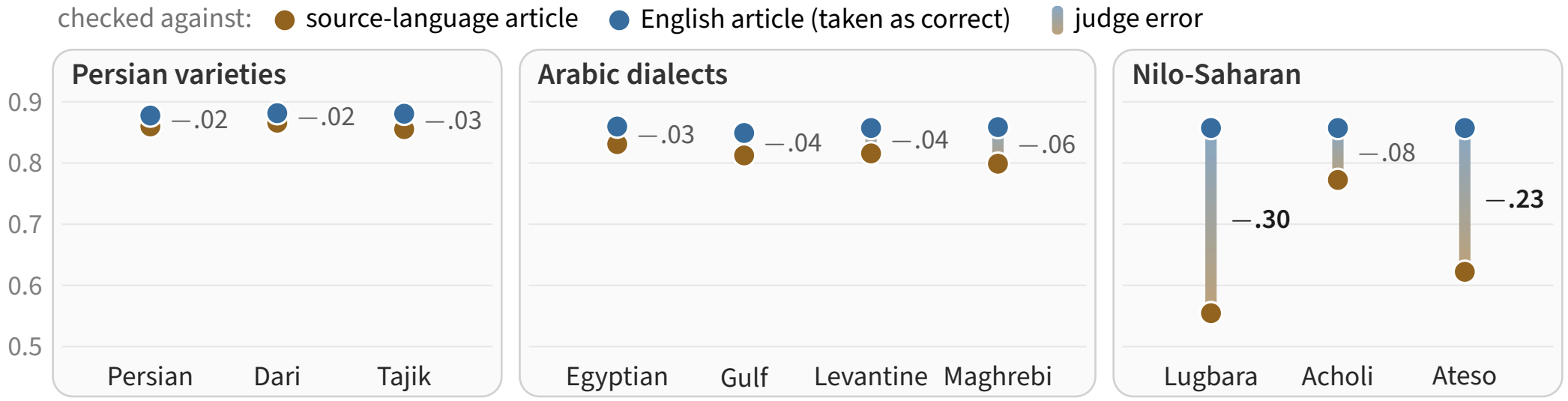
\begin{figure}[t]
\centering
\begin{tikzpicture}[x=1cm,y=1cm,font=\sffamily\scriptsize,every node/.style={inner sep=1pt},
  panel/.style={draw=black!18,line width=.5pt,rounded corners=5pt,fill=black!2},
  head/.style={font=\sffamily\bfseries\fontsize{7.6}{9}\selectfont,text=black!80},
  grid/.style={draw=black!9,line width=.4pt},
  gap/.style={bottom color=cbGold!60,top color=cbBlue!60,rounded corners=1.2pt},
  dotsrc/.style={circle,fill=cbGold,draw=white,line width=.6pt,inner sep=0pt,minimum size=5.6pt},
  doten/.style={circle,fill=cbBlue,draw=white,line width=.6pt,inner sep=0pt,minimum size=5.6pt}]
\draw[panel] (0.55,-0.44) rectangle (4.51,2.62);
\node[head,anchor=north west] at (0.67,2.55) {Persian varieties};
\draw[panel] (4.67,-0.44) rectangle (9.75,2.62);
\node[head,anchor=north west] at (4.79,2.55) {Arabic dialects};
\draw[panel] (9.91,-0.44) rectangle (13.87,2.62);
\node[head,anchor=north west] at (10.03,2.55) {Nilo-Saharan};
\draw[grid] (0.63,0.000) -- (4.43,0.000);
\draw[grid] (4.75,0.000) -- (9.67,0.000);
\draw[grid] (9.99,0.000) -- (13.79,0.000);
\node[anchor=east,text=black!50] at (0.50,0.000) {0.5};
\draw[grid] (0.63,0.540) -- (4.43,0.540);
\draw[grid] (4.75,0.540) -- (9.67,0.540);
\draw[grid] (9.99,0.540) -- (13.79,0.540);
\node[anchor=east,text=black!50] at (0.50,0.540) {0.6};
\draw[grid] (0.63,1.080) -- (4.43,1.080);
\draw[grid] (4.75,1.080) -- (9.67,1.080);
\draw[grid] (9.99,1.080) -- (13.79,1.080);
\node[anchor=east,text=black!50] at (0.50,1.080) {0.7};
\draw[grid] (0.63,1.620) -- (4.43,1.620);
\draw[grid] (4.75,1.620) -- (9.67,1.620);
\draw[grid] (9.99,1.620) -- (13.79,1.620);
\node[anchor=east,text=black!50] at (0.50,1.620) {0.8};
\draw[grid] (0.63,2.160) -- (4.43,2.160);
\draw[grid] (4.75,2.160) -- (9.67,2.160);
\draw[grid] (9.99,2.160) -- (13.79,2.160);
\node[anchor=east,text=black!50] at (0.50,2.160) {0.9};
\shade[gap] (1.360,1.943) rectangle (1.460,2.039);
\node[dotsrc] at (1.410,1.943) {};
\node[doten] at (1.410,2.039) {};
\node[anchor=west,text=black!65] at (1.530,1.991) {$-$.02};
\node[anchor=north,text=black!85] at (1.410,-0.130) {Persian};
\shade[gap] (2.480,1.976) rectangle (2.580,2.060);
\node[dotsrc] at (2.530,1.976) {};
\node[doten] at (2.530,2.060) {};
\node[anchor=west,text=black!65] at (2.650,2.018) {$-$.02};
\node[anchor=north,text=black!85] at (2.530,-0.130) {Dari};
\shade[gap] (3.600,1.919) rectangle (3.700,2.055);
\node[dotsrc] at (3.650,1.919) {};
\node[doten] at (3.650,2.055) {};
\node[anchor=west,text=black!65] at (3.770,1.987) {$-$.03};
\node[anchor=north,text=black!85] at (3.650,-0.130) {Tajik};
\shade[gap] (5.480,1.787) rectangle (5.580,1.942);
\node[dotsrc] at (5.530,1.787) {};
\node[doten] at (5.530,1.942) {};
\node[anchor=west,text=black!65] at (5.650,1.864) {$-$.03};
\node[anchor=north,text=black!85] at (5.530,-0.130) {Egyptian};
\shade[gap] (6.600,1.687) rectangle (6.700,1.885);
\node[dotsrc] at (6.650,1.687) {};
\node[doten] at (6.650,1.885) {};
\node[anchor=west,text=black!65] at (6.770,1.786) {$-$.04};
\node[anchor=north,text=black!85] at (6.650,-0.130) {Gulf};
\shade[gap] (7.720,1.705) rectangle (7.820,1.929);
\node[dotsrc] at (7.770,1.705) {};
\node[doten] at (7.770,1.929) {};
\node[anchor=west,text=black!65] at (7.890,1.817) {$-$.04};
\node[anchor=north,text=black!85] at (7.770,-0.130) {Levantine};
\shade[gap] (8.840,1.614) rectangle (8.940,1.937);
\node[dotsrc] at (8.890,1.614) {};
\node[doten] at (8.890,1.937) {};
\node[anchor=west,text=black!65] at (9.010,1.775) {$-$.06};
\node[anchor=north,text=black!85] at (8.890,-0.130) {Maghrebi};
\shade[gap] (10.720,0.295) rectangle (10.820,1.929);
\node[dotsrc] at (10.770,0.295) {};
\node[doten] at (10.770,1.929) {};
\node[anchor=west,font=\sffamily\bfseries\scriptsize,text=black!90] at (10.890,1.112) {$-$.30};
\node[anchor=north,text=black!85] at (10.770,-0.130) {Lugbara};
\shade[gap] (11.840,1.471) rectangle (11.940,1.929);
\node[dotsrc] at (11.890,1.471) {};
\node[doten] at (11.890,1.929) {};
\node[anchor=west,text=black!65] at (12.010,1.700) {$-$.08};
\node[anchor=north,text=black!85] at (11.890,-0.130) {Acholi};
\shade[gap] (12.960,0.660) rectangle (13.060,1.929);
\node[dotsrc] at (13.010,0.660) {};
\node[doten] at (13.010,1.929) {};
\node[anchor=west,font=\sffamily\bfseries\scriptsize,text=black!90] at (13.130,1.294) {$-$.23};
\node[anchor=north,text=black!85] at (13.010,-0.130) {Ateso};
\node[anchor=west,text=black!55] (l0) at (0.55,2.88) {checked against:};
\node[dotsrc] (l1) at ([xshift=7pt]l0.east) {};
\node[anchor=west] (l2) at ([xshift=1pt]l1.east) {source-language article};
\node[doten] (l3) at ([xshift=9pt]l2.east) {};
\node[anchor=west] (l4) at ([xshift=1pt]l3.east) {English article (taken as correct)};
\shade[gap] ([xshift=9pt,yshift=-3.6pt]l4.east) rectangle ([xshift=11.8pt,yshift=3.6pt]l4.east);
\node[anchor=west] at ([xshift=13.5pt]l4.east) {judge error};
\end{tikzpicture}%
\caption{English-input summaries scored by Claude Sonnet 4.6 against the source-language article (gold) and the English article (blue), averaged over the systems of Table~\ref{tab:2x2}.}
\label{fig:judge-bars}
\end{figure}

\paragraph{Details.}
We evaluate Claude Sonnet 4.6 as the Judge language model, with mean faithfulness scores calculated over the same systems that we evaluated faithfulness on: GPT-5.5, gpt-oss-20B, Qwen3-30B-A3B, Qwen3-8B and Aya-23-8B, plus the specialists Dorna2-8B (Persian varieties), Fanar-1-9B (Arabic dialects) and both Sunflower models (Nilo-Saharan languages).
\paragraph{Results.} In Figure~\ref{fig:judge-agree}, we plot claim agreement between the LLM judge when evaluating against the English and low-resource source document. Claim agreement is highest for Persian varieties, followed by Arabic dialects. Nilo-Saharan languages have low agreement, with a Cohen's $\kappa$ between 0.05 and 0.18. In Figure~\ref{fig:judge-bars}, we additionally compare the differences in average faithfulness scores across source document language. Notably, when evaluating against the low-resource source document, the LLM-as-a-Judge approach rates summaries of Nilo-Saharan documents as significantly less faithful. The difference is most pronounced for Lugbara (-0.30). Overall, these results show that LLM-as-a-Judge approaches that are validated in high-resource settings can fail to generalize to low-resource language reference documents.

\section{Discussion}

There is scarcity of benchmarks for evaluating the multilingual summarization capabilities of language models from low-resource languages and dialects. Using \gls{approach}, we automatically crafted a cross-lingual summarization benchmark of paired low-resource and high-resource documents that covered ten source languages. We utilize the benchmark to evaluate language model summarization faithfulness and LLM-as-Judge performance.

\textbf{RQ1: How well do LLMs summarize low-resource languages?} Summarization quality is inconsistent across languages, and \textit{large gaps can exist between low-resource and high-resource settings}. The frontier model we evaluated, GPT-5.5, outperformed both open-source and specialist models in many settings. However, lower-resource Nilo-Saharan languages were an exception where specialist Sunflower-32B and Sunflower-9B models were able to outperform GPT-5.5 in faithfulness. The strongest frontier model loses much of its faithfulness when the input article is in a Nilo-Saharan language: GPT-5.5 scores 0.90 with English input, but 0.47 with Lugbara, 0.74 with Acholi and 0.62 with Ateso input. Summarization from very low-resource languages therefore remains unsolved even for frontier models. Without a parallel English article this loss would be hard to measure, since the judge cannot reliably verify claims against the Lugbara or Ateso article.

\textbf{RQ2: Does LLM-as-a-Judge Performance Generalize to Low-Resource Languages?} In our experiments, \textit{a strong LLM-as-a-Judge can fail} for certain languages. Sonnet 4.6 had high agreement with native-speaker annotators when comparing summaries to English documents. However, its behavior degraded substantially when verifying claims against Nilo-Saharan reference documents.

Overall, our work identifies gaps in model performance across languages in current language models, as well as highlights ways that existing datasets can be repurposed to elucidate these limitations.

\section{Limitations}
\label{sec:limitations}
Our approach has several limitations. First, the quality of the benchmark depends on the register of its source data. The Arabic articles are built from travel phrasebooks and conversational sentences, so they test comprehension of conversational dialect and read less like documents than real news does. The Nilo-Saharan articles, built from written prose, read much closer to real documents (Appendix~\ref{app:qualitative}), which suggests that the gap comes from the source data rather than the pipeline. Second, the English side is only as good as its human translations. Third, the dialect labels are the regional groups of AraBench, so Gulf also covers Iraqi and Yemeni cities and Egyptian covers Sudanese ones; every article is released with its country code for finer analysis. Finally, our pipeline's performance is based on many factors from composer choice to judge choice, every model has to be chosen carefully, and so does the cluster size relative to the size of the dataset.

\subsection*{AI Use Statement}
AI was used for assistance with literature review, polishing the manuscript, and for agentic coding. Additionally language models were employed for synthetic data generation.

\subsection*{Ethics Statement}
The native-speaker study was reviewed and approved by the Columbia University IRB. Annotators were recruited through Prolific and Upwork with the same screening questionnaire, gave informed consent, and were paid 17 USD per hour for a session of about one hour. Responses are reported only in aggregate. About half a percent of articles were manually identified as containing charged material inherited from the source corpus, which were excluded from the survey. The Arabic sentence pool is the public text release of AraBench (MADAR and QAraC credited), and the ten articles that contain sentences of the research-only MDC corpus are withheld; the Persian pools derive from FLORES and the Lugbara pool from SALT, each released under its own terms with attribution (Appendix~\ref{app:licensing}).

\subsection*{Reproducibility Statement}
The pipeline code, the exact prompts (Appendix~\ref{app:prompts}), the benchmark, and all results and scripts are available at \url{https://github.com/Ajaberr/ParaWeave}. Model versions and settings are in Appendix~\ref{app:models}. Construction depends on API models and is not bit-reproducible, so the full construction log is released for re-auditing.

\paperacknowledgments

\bibliography{references}
\bibliographystyle{iclr2027_conference}

\appendix
\section{Human Study Detail}
\label{app:human}
Each dialect has its own survey of 58 items: 8 odd-one-out items, 17 checklist articles, 16 articles for the two-pass ratings (4 per hidden condition), and 17 faithfulness claims. Each annotator is a native speaker of one dialect, screened with an eligibility questionnaire, and works on that dialect only for about one hour. Annotators were picked by the region they grew up in and their self-reported native dialect; for Gulf no such annotator was available, so we used annotators from other regions who reported Gulf proficiency and confirmed they understood the articles.

\paragraph{The four tasks.}
Every annotator completed four tasks on articles in their own dialect: the first three validate the benchmark (Section~\ref{sec:validation}) and the fourth the LLM judge (Section~\ref{sec:rq3}); the instructions, question wording and answer options follow under Instrument.
\begin{enumerate}
\item \emph{Odd one out.} Five sentences, four from one article and one from another, and the annotator picks the one that does not belong, an intrusion test inspired by \citet{chang2009tealeaves} and \citet{shen2021coherence}. This checks that articles are distinct enough for a native speaker to tell apart. This is to validate article grouping.
\item \emph{Quality checklist.} For an accepted article, the annotator ticks every quality issue that applies from a fixed list (repetition, unclear reference, grammaticality, focus, lacking sensible meaning, cut off, other) or none; the first four are the DUC quality questions \citep{dang2005duc}, as used by SummEval \citep{fabbri2021summeval}. This is to evaluate overall article quality.
\item \emph{Coherence and human-likeness ratings.} Sixteen articles from four conditions that the annotator is not told about: natural news, accepted articles, articles the gate rejected, and accepted articles with their sentence order destroyed by k-block shuffling \citep{barzilay2008entity,laban2021shuffle}. Each article is rated twice on a four-point scale, once for ``How coherent is this article?'' and once for ``How human-written does this article sound?'' (inspired by \citealp{ippolito2020detection}); a rating of 1 or 2 asks for the reasons from a checklist. This is to evaluate the gating mechanism.
\item \emph{Claim faithfulness.} The annotator sees the Arabic source article and one English claim taken from a system summary and rates how well the source supports it, from fully supported to fully contradicted \citep{min2023factscore}, naming the error type for anything below fully supported \citep{pagnoni2021frank,alansari2025arahallueval}. This is the task that validates the LLM judge of Section~\ref{sec:rq3}.
\end{enumerate}

\paragraph{Instrument.}
The survey was run in Qualtrics; the wording below is reproduced from the survey file. The welcome page told annotators that the articles are ``a mix of two kinds: some were written entirely by a human, and some were assembled by an AI system out of real human-written sentences (the AI selects and orders existing sentences; it does not invent new ones). You will not be told which is which.''
\begin{itemize}
\item \textbf{Odd one out}: ``Which ONE of these five sentences does not belong with the others?''
\item \textbf{Quality checklist}: ``Read the article below, then select EVERY quality issue that applies to it from the checklist underneath. You may select as many issues as apply, or none at all if the article has no such issues.'' Options: unnecessary repetition; referential clarity issue; grammaticality problem (bad enough to make the meaning unclear); focus issue; lacks sensible meaning; incomplete or feels cut off; other; none of the above.
\item \textbf{Coherence}: ``How coherent is this article?'' with 1 not at all coherent, 2 somewhat coherent, 3 mostly coherent, 4 completely coherent. For a rating of 1 or 2, the follow-up asks for every type of incoherence observed: topic scatter; shared setting only / phrasebook (sentences share a domain or setting but are independent, one-off lines, not a continuous scene); broken order or structure; contradictory or inconsistent content.
\item \textbf{Human-written}: ``How human-written does this article sound?'' with 1 does not sound human-written at all, 2 somewhat, 3 mostly, 4 completely. For a rating of 1 or 2, the follow-up lists: repetitive or redundant phrasing; mismatched voice, register or formality; generic or formulaic phrasing; uneven level of detail or pacing.
\item \textbf{Claim faithfulness}: the Arabic source article is shown above an English claim, then ``How well is this claim supported by the source? Rate from Fully supported to Fully contradicted. Judge the claim against the Arabic source above.'' Options: (2) fully supported, (1) partially supported, (0) irrelevant / not supported, ($-1$) partially contradicted, ($-2$) fully contradicted. Anything below fully supported asks for one error type: entity, number, relation, unsupported, overgeneralization, other. For comparison with the judge's four labels, the two contradicted levels are merged.
\end{itemize}

\paragraph{Gate against the annotators.}
On the 32 accepted and rejected articles of task 3, the gate agrees with the annotators' mean coherence rating (coherent at 3 or above) on 26 articles ($\kappa=0.62$), while two annotators of the same dialect agree with each other at $\kappa=0.13$. Table~\ref{tab:reasons} gives the reasons for low ratings.

\begin{table}[H]
\centering
\small
\begin{tabular}{lccccc}
\toprule
\textbf{Condition} & \textbf{Rated 1--2} & \textbf{Broken order} & \textbf{Shared setting only} & \textbf{Topic scatter} & \textbf{Contradictory} \\
\midrule
Natural news & 0/28 & 0 & 0 & 0 & 0 \\
Accepted & 9/28 & 9 & 3 & 1 & 3 \\
Rejected & 21/28 & 18 & 13 & 4 & 4 \\
Shuffled & 22/28 & 20 & 9 & 3 & 7 \\
\bottomrule
\end{tabular}
\caption{Reasons given for coherence ratings of 1 or 2 (several allowed; 28 ratings per condition). Rejected articles are faulted for ``shared setting only'' far more often than accepted ones.}
\label{tab:reasons}
\end{table}

\paragraph{Judge against the annotators.}
On whether a claim is fully supported, the judge and the annotators agree on 70\% of claims ($\kappa=0.40$), and two annotators agree on 73\% ($\kappa=0.45$) of the 51 claims both rated. The claims were drawn from those on which the judge's verdicts against the Arabic and English articles agree, so this task cannot adjudicate the disagreements of Section~\ref{sec:rq2}.

\section{Prompts}
\label{app:prompts}
The prompts below are reproduced from the released code without change; placeholders in double underscores or braces are filled per call, and the source-language name is a parameter (``dialectal Arabic'' for the Arabic build). System prompt first, then the user message.
\paragraph{Composer.}
\begin{lstlisting}
You are given a topical CLUSTER of dialectal Arabic sentences with English translations (numbered SOURCE ||| EN pairs). Select and order sentences into ARTICLES that each score >=4 on this rubric.
STRICT COHERENCE RUBRIC (the ONLY thing that counts):
  5 = ONE topic/event/scene: the SAME participants and a single continuous through-line (one conversation, one incident, one story).
  4 = one scene with only minor drift.
  3 = loosely related: a shared SETTING/DOMAIN but DIFFERENT sub-things -- e.g. several separate accident-related interactions, or a list of payment/shopping/tourism phrases. A shared setting/domain/topic is NOT coherence. REJECT (do NOT output).
  1-2 = grab-bag / phrasebook / independent one-off utterances (tourism Qs, food orders, symptom lists, single-word commands).
Each article = 5+ pairs forming a GENUINE single connected scene/story (rubric >=4). ALLOWED edits ONLY: (1) reorder pairs for flow; (2) unify PROPER-NOUN names (same change to BOTH the dialectal Arabic and the English). FORBIDDEN: rewording, adding connectives, changing common nouns/verbs, adding/removing content. Be STRICT: only output articles you are confident are >=4 (a single connected scene). A cluster that is only a shared setting yields NOTHING -- return []. If no single connected scene exists, return []. Return ONLY JSON: {"articles":[{"pairs":[{"ar":"..","en":".."}, ...]}, ...]}.

<the cluster as numbered lines: i. SOURCE ||| EN>
\end{lstlisting}
\paragraph{Composer, rebuild after a rejected candidate (user message; same system prompt).}
\begin{lstlisting}
This article was judged loosely coherent. These sentence indices do not belong: __DROP__. Rebuild ONE tighter article as a connected subset (drop those + anything else off-scene; reorder allowed; NO rewording/new content). Keep >=5 pairs or return empty. Numbered SOURCE ||| EN pairs:
__BODY__

Return ONLY JSON {"pairs":[{"ar":"..","en":".."}, ...]}.
\end{lstlisting}
\paragraph{Gate, coherence check.}
\begin{lstlisting}
You rate the COHERENCE of a candidate article on this exact scale. Judge STRICTLY; a shared setting/domain is a 3, NOT a 4.
STRICT COHERENCE RUBRIC (the ONLY thing that counts):
  5 = ONE topic/event/scene: the SAME participants and a single continuous through-line (one conversation, one incident, one story).
  4 = one scene with only minor drift.
  3 = loosely related: a shared SETTING/DOMAIN but DIFFERENT sub-things -- e.g. several separate accident-related interactions, or a list of payment/shopping/tourism phrases. A shared setting/domain/topic is NOT coherence. REJECT (do NOT output).
  1-2 = grab-bag / phrasebook / independent one-off utterances (tourism Qs, food orders, symptom lists, single-word commands).

Numbered English sentences:
__BODY__

Return ONLY JSON {"coherence": <1-5 per the rubric>, "drop":[<indices NOT part of the single connected scene>], "reason":"<=15 words"}.
\end{lstlisting}

\paragraph{Summarizer (identical for every evaluated system; maxw = 30).}
\begin{lstlisting}
You are a precise cross-lingual summarizer. You read Arabic (including dialect) and write English.

Summarize the following {dialect}-Arabic text in ONE English sentence of at most {maxw} words. Output ONLY the summary as plain declarative English -- no preamble, no quotes, no commentary. Do not add any fact that is not stated in the text.

TEXT:
{src}
\end{lstlisting}
\paragraph{Claim decomposition.}
\begin{lstlisting}
You break a summary into atomic facts. An atomic fact states ONE thing, is fully self-contained (resolve every pronoun to an explicit entity), and copies no opinion or hedge. Output ONLY facts that the summary asserts. Many summaries are unattributed first-person dialogue with no named speaker -- in that case resolve 'I'/'me' to 'the speaker' and 'you' to 'the listener' (or 'the addressee'); do NOT return zero claims just because no proper name is given.

Decompose this summary into atomic factual claims. Return ONLY JSON: {"claims": ["<claim 1>", "<claim 2>", ...]}. Each claim is one short standalone English sentence with all references made explicit. If the summary states nothing factual, return {"claims": []}.

SUMMARY:
<summary text>
\end{lstlisting}
\paragraph{Claim judging.}
\begin{lstlisting}
You are a strict cross-lingual faithfulness judge. You read Arabic (including dialect) and English. Given a SOURCE text and CLAIMS taken from an English summary of it, decide for EACH claim whether the SOURCE supports it. Judge ONLY against the source; use no outside knowledge.

SOURCE:
__SRC__

CLAIMS:
__CLAIMS__

For each claim return a verdict. Labels: SUPPORTED (fully stated/entailed by the source), PARTIAL (partly supported, some detail missing or vague), NOT_SUPPORTED (the source does not state it -- fabricated/extrinsic), CONTRADICTED (the source states the opposite). error in { none | entity: wrong/invented named thing (person/place/org) | number: wrong quantity/date/amount/count | relation: right entities but wrong who-did-what/attribution/causality | unsupported: content absent from the source (fabrication) | contradiction: asserts the opposite of the source | overgeneralization: broader/stronger than the source supports }. Return ONLY JSON: {"verdicts":[{"i":1,"label":"...","error":"...","evidence":"<short source quote or empty>","reason":"<=15 words"}]} with one entry per claim index.
\end{lstlisting}
\paragraph{Translation probe (Appendix~\ref{app:translation}; greedy decoding).}
\begin{lstlisting}
You are a precise translator. Translate the given __LANG__ sentence into English. Output ONLY the English translation as plain text -- no preamble, no quotes, no commentary, no transliteration.

__LANG__:
__SENTENCE__

English:
\end{lstlisting}

\section{Models and Settings}
\label{app:models}
\begin{table}[H]
\centering
\small
\setlength{\tabcolsep}{4.5pt}
\begin{tabular}{llrrrrr}
\toprule
\textbf{Variety} & \textbf{Source pairs} & \textbf{Pool} & \textbf{Articles} & \textbf{Sentences} & \textbf{Sent./art.} & \textbf{En words/art.} \\
\midrule
Egyptian & MADAR/BTEC & 17{,}236 & 301 & 2{,}397 & 8.0 & 69 \\
Gulf & MADAR/BTEC & 36{,}045 & 914 & 7{,}656 & 8.4 & 82 \\
Levantine & MADAR/BTEC & 21{,}507 & 435 & 3{,}481 & 8.0 & 70 \\
Maghrebi & MADAR/BTEC & 30{,}802 & 680 & 5{,}406 & 8.0 & 69 \\
\textbf{Arabic dialects} & & \textbf{105{,}590} & \textbf{2{,}330} & \textbf{18{,}940} & \textbf{8.1} & \textbf{74} \\
\midrule
Persian & FLORES-200 & 2{,}009 & 232 & 997 & 4.3 & 91 \\
Dari & FLORES-200 & 2{,}009 & 230 & 1{,}001 & 4.4 & 91 \\
Tajik & FLORES-200 & 2{,}009 & 159 & 701 & 4.4 & 93 \\
\midrule
Lugbara & SALT & 24{,}946 & 240 & 3{,}571 & 14.9 & 142 \\
Acholi & SALT & -- & 240 & 3{,}571 & 14.9 & 142 \\
Ateso & SALT & -- & 240 & 3{,}571 & 14.9 & 142 \\
\bottomrule
\end{tabular}
\caption{Composition of every article set and the size of the sentence pool it was built from; the Acholi and Ateso articles are the Lugbara articles in those languages.}
\label{tab:arabic-instantiation}
\end{table}

\begin{itemize}
\item \textbf{Embedding.} Qwen3-Embedding-8B (\texttt{Qwen/Qwen3-Embedding-8B}, sentence-transformers, float16, batch 16) embeds the English side under the instruction ``Instruct: Group sentences that belong to the same conversation, scene, or connected event.\textbackslash nQuery:~'' and the 4096-d vectors are L2-normalised. Each round reduces the surviving pool with scikit-learn \texttt{PCA} to 50 dimensions and clusters it per sub-dialect with \texttt{MiniBatchKMeans} ($k=\lfloor n/150\rfloor$, batch 512, \texttt{n\_init=auto}), both with \texttt{random\_state=0}.
\item \textbf{Composition.} GPT-5.5, one cluster per call. The model is \texttt{gpt-5.5-2026-04-23} through the OpenAI API in JSON mode with \texttt{reasoning\_effort=low} and \texttt{max\_completion\_tokens=4096}; GPT-5 models fix temperature at the API default. At most 250 sentence pairs of a cluster are shown, every article must have at least five pairs, and a candidate the gate rejects may be rebuilt at most twice from the gate's list of off-scene sentences. The released Arabic articles were composed in July 2026 (final build 2026-07-08).
\item \textbf{Gate.} A separate frontier model (Claude Sonnet 4.6). It runs as \texttt{claude-sonnet-4-6} through the Messages API at temperature 0 in two calls per candidate: a coherence check on the English side that returns a 1--5 score and the indices of off-scene sentences (\texttt{max\_tokens=300}), then the four-axis rubric on both sides (\texttt{max\_tokens=150}); a candidate is kept when the rubric coherence is at least 4 and it has at least 7 sentences. A build runs for at most 12 rounds (the Arabic dialects) or 6 (the FLORES varieties) and stops earlier when a round keeps fewer than 5 articles (2 for FLORES); the Lugbara build ran 3 rounds and was then resumed for 10 more, 13 in all (Table~\ref{tab:recycling}).
\item \textbf{Evaluated systems.} GPT-5.5 (\texttt{gpt-5.5-2026-04-23}), gpt-oss-20B, Qwen3-30B-A3B, Qwen3-8B, Fanar-1-9B-Instruct, Aya-23-8B. The open-weight systems are \texttt{openai/gpt-oss-20b}, \texttt{Qwen/Qwen3-30B-A3B}, \texttt{Qwen/Qwen3-8B}, \texttt{QCRI/Fanar-1-9B-Instruct} and \texttt{CohereForAI/aya-23-8B}, run in-process with transformers (\texttt{torch\_dtype=auto}) on one GPU; GPT-5.5 runs through the OpenAI API. Every system receives the same prompt (Appendix~\ref{app:prompts}) asking for one English sentence of at most 30 words, decoded with temperature 0.3, top-$p$ 0.8 and at most 90 new tokens; GPT-5.5 uses \texttt{reasoning\_effort=low} at its default temperature, Qwen3 runs with thinking off, and gpt-oss runs at low reasoning effort with only its final channel kept. Summaries were generated on 2026-07-15 and 2026-07-16.
\item \textbf{Judge.} Claude Sonnet 4.6 (\texttt{claude-sonnet-4-6}) through the Batches API, temperature 0. Each call judges all claims of one summary against one reference with \texttt{max\_tokens=900} and the source truncated to 8{,}000 characters; the full run is about 105{,}700 calls made between 2026-07-15 and 2026-07-17. Claims are extracted once per summary by the same model with the decomposition prompt (temperature 0, \texttt{max\_tokens=400}) and frozen, so both references are checked on identical claims; a live re-run of the decomposer recovered 5{,}537 of the 5{,}995 summaries for which the first pass returned no claims. Judged summaries exist for 2{,}170 of the 2{,}330 released articles, and the result tables are computed on those.
\item \textbf{Transfer runs.} The Acholi and Ateso versions of the Lugbara articles were summarized on 2026-09-25 with every system's Lugbara settings and judged by the same judge; their multi-way SALT text is read from the text columns of \texttt{mekaneeky/Synthetic\_Acholi\_MMS}, an ungated release that carries SALT's text table. Lugbara and the Persian varieties (Section~\ref{sec:transfer}, Appendix~\ref{app:transfer}) use the same composer, gate, prompts and judge with the language named in place of the dialect; the Lugbara build ran with a minimum of 12 sentences per article and the FLORES builds with a minimum of 4 (their sentences average 21 English words against 8 for MADAR), both in September 2026. The added systems are \texttt{Sunbird/Sunflower-Qwen3.5-9B} (Qwen3.5-9B with continued pretraining, instruction tuning and GRPO on Ugandan and other African-language data, Lugbara in its middle coverage tier; preview weights, revision \texttt{3537737}, downloaded 2026-09-21, served in bf16 by vLLM with thinking off and greedy decoding; its card does not list training corpora but evaluates on the SALT test split), \texttt{Sunbird/Sunflower-32B} (Qwen3-32B fine-tuned on 31 Ugandan languages, whose training mix includes SALT) and \texttt{PartAI/Dorna2-Llama3.1-8B-Instruct}. Sunflower-32B is loaded in 4-bit NF4 with bitsandbytes, and for these sets Qwen3-30B-A3B is served by vLLM with fp8 weight-only quantisation, because neither fits one of our 48\,GB GPUs in bf16 and sharded bf16 inference on the shared machine produced degenerate output; both decode greedily, every other system as in the Arabic run. Summaries whose decomposition yields no claim are excluded from the means; with Lugbara input there are 120 such summaries for gpt-oss-20B, 57 for Qwen3-8B, 30 for Qwen3-30B-A3B and at most 7 for any other system. The reasoning-effort sweep of Figure~\ref{fig:effort} runs \texttt{openai.gpt-5.6-sol} through Bedrock Mantle at each effort it accepts (none, low, medium, high and xhigh; minimal is rejected) with \texttt{max\_completion\_tokens=32000} and the summarizer prompt unchanged, on 2026-09-23 and 2026-09-24.
\item \textbf{Dialect specialists, excluded.} Nile-Chat-12B (Egyptian; \citealp{shang2025nilechat}) and Atlas-Chat-9B (Moroccan; \citealp{shang2024atlaschat}) were run on an earlier article cohort and excluded because they return Arabic rather than an English summary.
\item \textbf{Seeds.} PCA and KMeans use \texttt{random\_state=0}; bootstrap confidence intervals and the paired-bootstrap tiers use 1{,}000 to 2{,}000 resamples with seed 0; the survey items were drawn with seed 20260909, and the shuffled condition moves blocks of $k=\max(2,\lfloor n/4\rfloor)$ sentences, redrawn until at least 60\% of the sentences change position.
\item \textbf{Anchors and metrics.} AlignScore-large checkpoint \citep{zha2023alignscore} in \texttt{nli\_sp} mode; \texttt{MoritzLaurer/mDeBERTa-v3-base-mnli-xnli} \citep{he2023debertav3,laurer2024less}; MiniCheck \texttt{flan-t5-large} \citep{tang2024minicheck}; \texttt{intfloat/multilingual-e5-large-instruct} \citep{wang2024me5}; chrF \citep{popovic2015chrf} from sacrebleu. The translation probe runs every open-weight system through vLLM 0.21 on one GPU in bfloat16 with greedy decoding (Qwen3-30B-A3B in fp8, as for its summaries; Sunflower-32B in 4-bit NF4 through transformers); the Arabic translations of gpt-oss-20B, Qwen3-8B, Fanar-1-9B and Aya-23-8B come from an earlier run of the same prompt through transformers.
\end{itemize}

We reduce the 4096-dimensional representation to $d=50$ with PCA \citep{jolliffe2016pca} before clustering, motivated by the connection between the principal components and a continuous relaxation of the K-means objective \citep{ding2004kmeans}.

The size of the clusters can be adjusted according to the amount of parallel data available and the context window of the composer model to be used.

\section{Construction Ablation Detail}
\label{app:ablation}
\paragraph{High-resource clustering.}
Our pipeline clusters the corresponding English side rather than the low-resource
sentence itself. We evaluate this choice only on real articles, whose boundaries nobody in this project chose, where positive pairs come from the same article and negatives come from different articles, and measure
whether compatible pairs rank above negatives:
\begin{equation}
\mathrm{AUC}
=
\Pr\left( s(p^{+}) > s(p^{-}) \right).
\end{equation}
Here $s$ is the cosine similarity between the embeddings of the two sentences of a pair $p=(i,j)$,
\begin{equation}
e_i=f(t_i;\iota),\qquad s(p)=\cos(e_i,e_j),
\end{equation}
so a configuration fixes the side $t_i$ that is embedded (the low-resource language or English) and the instruction $\iota$ (none or the scene instruction of Appendix~\ref{app:models}), and character $n$-gram TF-IDF on the same pairs gives a lexical baseline \citep{salton1988term}. The articles are FLORES-200 pages (Appendix~\ref{app:flores}), which every language shares, so each side is scored on identical pairs with each fragment's ten most similar fragments in English as hard negatives, and BBC articles from CrossSum \citep{bhattacharjee2023crosssum} in English and MSA, with hard negatives mined within each set.

On FLORES the low-resource embeddings reach $0.84$ to $0.86$ AUC on the Persian varieties and $0.86$ to $0.90$ on the eight Arabic dialects, above character TF-IDF at $0.73$ to $0.78$, but the English side reaches $0.92$ on the same pairs (Table~\ref{tab:auc}). We therefore cluster in English and transfer the resulting indices back to the corresponding low-resource sentences. This makes sense because smaller embedding models barely encounter these low-resource languages compared to the more high-resource languages.

\paragraph{Instruction conditioning.}
Prompting the English embedding model on a scenic instruction
raises $\mathrm{AUC}$ from $0.924$ to $0.943$ on FLORES and from $0.722$ to $0.734$ on CrossSum, and ties the uninstructed embedding on CrossSum MSA ($0.716$ against $0.719$), within the run-to-run variation of about $0.01$.

\begin{table}[H]
\centering
\small
\begin{tabular}{llccc}
\toprule
\textbf{Articles} & \textbf{Side} & \textbf{TF-IDF} & \textbf{Embedding} & \textbf{+ scene instruction} \\
\midrule
FLORES pages & English & 0.728 & 0.924 & 0.943 \\
 & Persian & 0.751 & 0.862 & 0.883 \\
 & Dari & 0.777 & 0.857 & 0.876 \\
 & Tajik & 0.735 & 0.841 & 0.870 \\
 & MSA Arabic & 0.749 & 0.900 & 0.907 \\
 & Arabic dialects (mean of 8) & 0.749 & 0.888 & 0.897 \\
\midrule
CrossSum (BBC) & English & 0.584 & 0.722 & 0.734 \\
 & MSA Arabic & 0.656 & 0.719 & 0.716 \\
\bottomrule
\end{tabular}
\caption{AUC of same-article sentence pairs on real articles (Qwen3-Embedding-8B). On FLORES every language is scored on the same pairs, and English beats every low-resource side.}
\label{tab:auc}
\end{table}

\section{Transfer Results: Other Language Families}
\label{app:transfer}
The transfer runs use the Arabic composer, gate, prompts and judge with the language name swapped in (settings and quantisation in Appendix~\ref{app:models}); the Lugbara build kept 240 articles of 12 to 29 sentences (mean 14.9) from 13 rounds over 24{,}946 pairs at a cost of \$79.95, and the FLORES sets are the recovered fragments of Appendix~\ref{app:flores}.

\begin{table}[H]
\centering
\small
\begin{tabular}{lcccc}
\toprule
\textbf{Variety} & \textbf{Round 1} & \textbf{All rounds} & \textbf{Rounds} & \textbf{All / round 1} \\
\midrule
Levantine & 82 & 435 & 12 & 5.3$\times$ \\
\midrule
Lugbara & 57 & 240 & 13 & 4.2$\times$ \\
Persian & 85 & 232 & 7 & 2.7$\times$ \\
Dari & 100 & 230 & 6 & 2.3$\times$ \\
Tajik & 53 & 159 & 6 & 3.0$\times$ \\
\bottomrule
\end{tabular}
\caption{Accepted articles after the first round and after the last. The first rounds of the Egyptian, Gulf and Maghrebi builds were not logged.}
\label{tab:recycling}
\end{table}

Summaries from English score about 0.86 against English but about 0.56 against Lugbara for every system (Table~\ref{tab:2x2}), so only the English-reference score measures the summariser, and the Sunflower models' margins over GPT-5.5 with Lugbara input (0.03 for the 9B, 95\% interval $-0.02$ to $0.08$; 0.14 for the 32B, $0.09$ to $0.18$) come with a caveat, since SALT is in the 32B's training mix and the 9B card evaluates on it. Against the Lugbara article the judge passes 76\% and 82\% of the false Lugbara-input claims of GPT-5.5 and gpt-oss-20B but 38 to 48\% of Aya-23-8B's and the Sunflower models', and rejects 29 to 31\% of the Sunflower models' true claims against 10 to 12\% of the GPT family's.

\paragraph{Acholi and Ateso.}
SALT is multi-way parallel, so replacing every sentence of the 240 Lugbara articles by its Acholi or Ateso translation gives articles with the identical English side (all 3{,}571 sentences map), which every system summarized with its Lugbara settings. The Sunflower models stay near $0.65$ on Acholi and Ateso, level with GPT-5.5 on Ateso (9B $+0.04$, 95\% interval $0.00$ to $0.09$) and below it on Acholi, perhaps because Acholi is a Luo language close to Dholuo, which FLORES-200 covers.

\begin{figure}[H]
\centering
\input{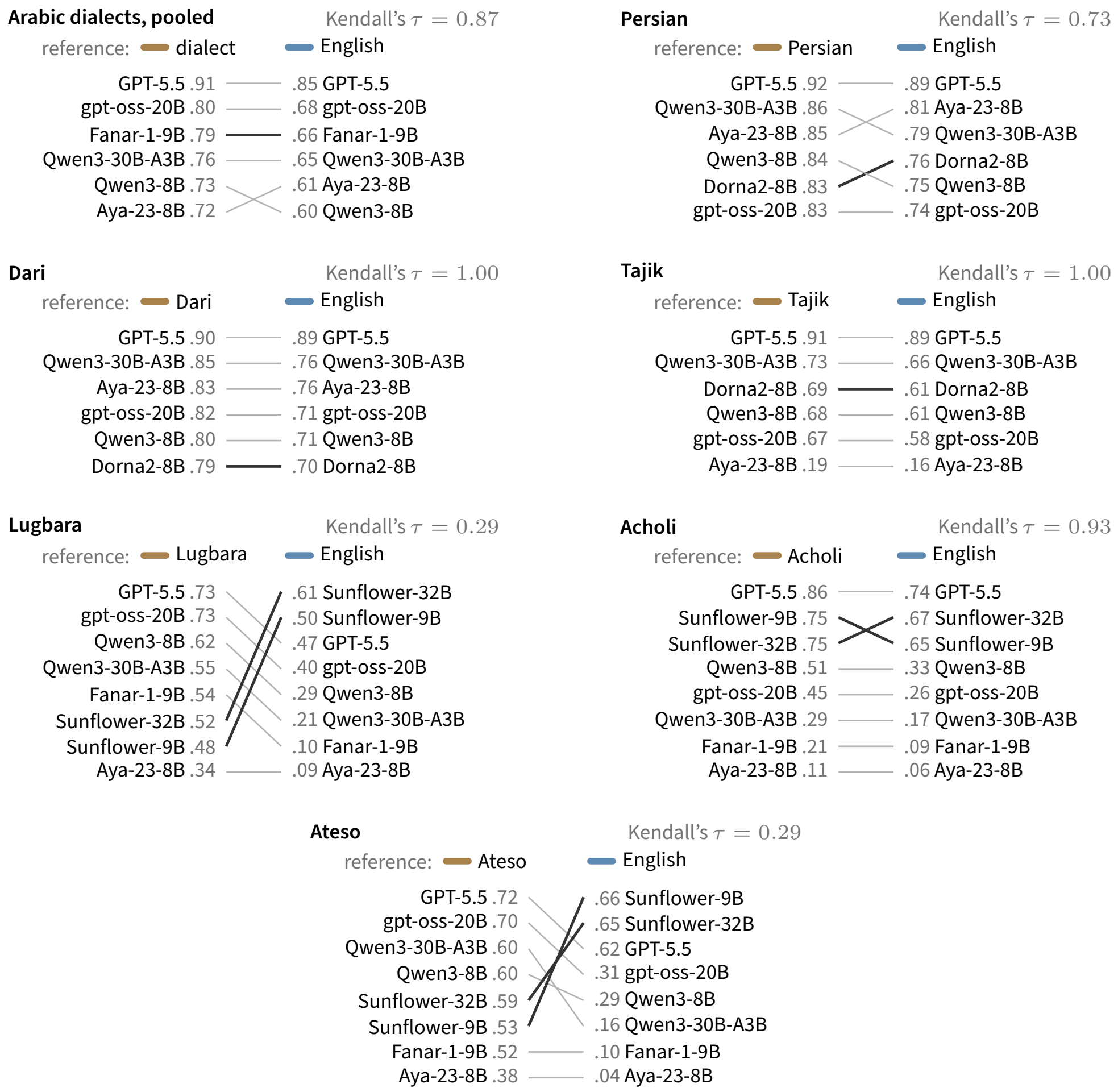}
\caption{Source-input summaries ranked against the source-language article (left) and its English version (right) in every build; on Lugbara the Sunflower models move from sixth and seventh to first and second.}
\label{fig:rankflip}
\end{figure}

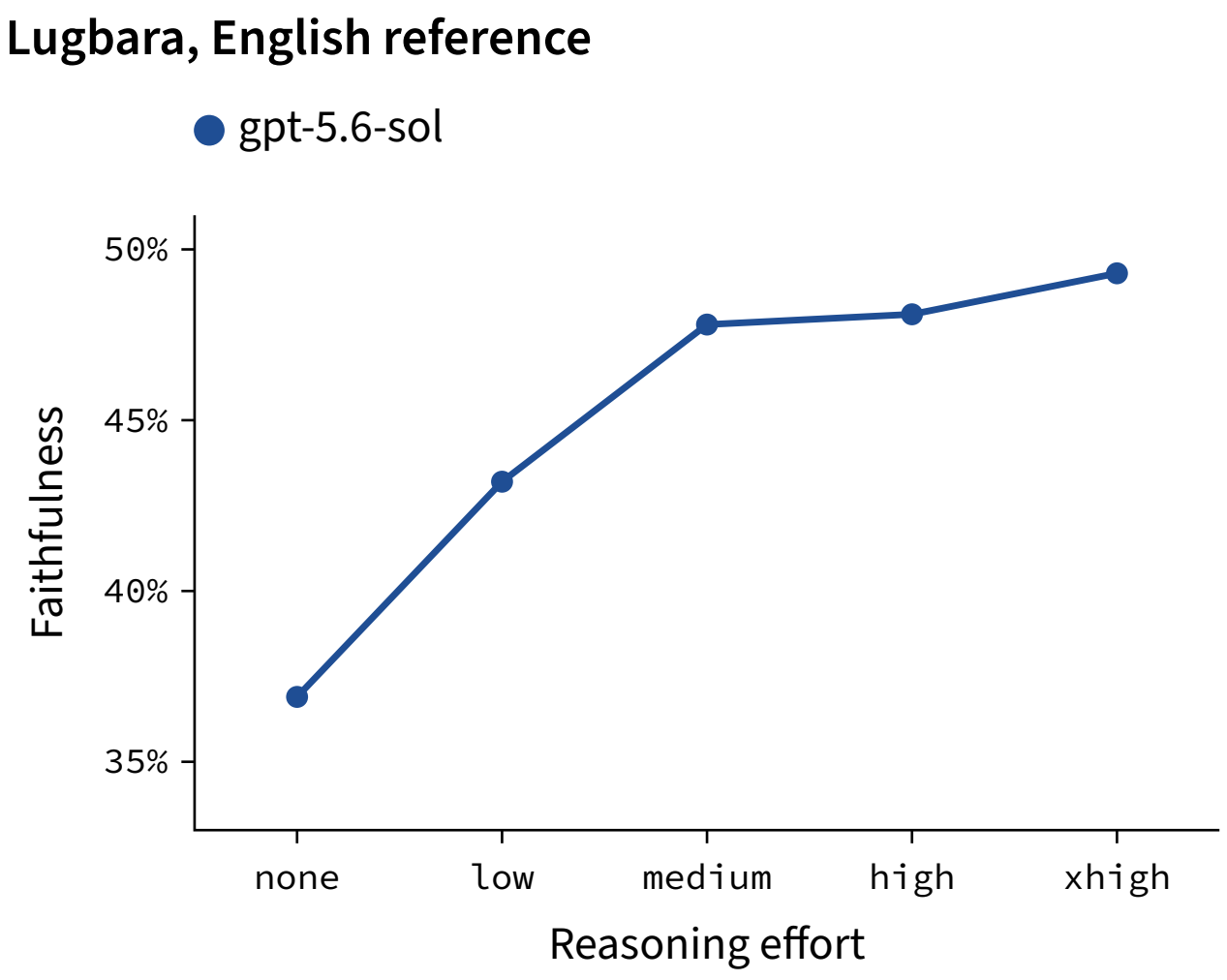
\begin{figure}[H]
\centering
\begin{tikzpicture}[x=1cm,y=1cm,font=\sffamily,
  axis/.style={draw=black,line width=0.5pt},
  tk/.style={draw=black,line width=0.5pt},
  tick/.style={font=\chartmono\scriptsize,text=black,inner sep=1.5pt}]
\path[use as bounding box] (-1.35,-1.05) rectangle (7.15,5.62);
\node[anchor=base west,font=\sffamily\bfseries\normalsize,inner sep=0pt] at (-1.30,5.30) {Lugbara, English reference};
\fill[oaiNavy] (0.10,4.78) circle (0.105);
\node[anchor=west,font=\sffamily\footnotesize,inner sep=0pt] at (0.29,4.78) {gpt-5.6-sol};
\draw[axis] (0,4.2) -- (0,0) -- (7.0,0);
\foreach \x/\l in {0.70/none, 2.10/low, 3.50/medium, 4.90/high, 6.30/xhigh}
  { \draw[tk] (\x,0) -- (\x,-0.09); \node[tick,anchor=base] at (\x,-0.40) {\l}; }
\foreach \y/\l in {0.4667/35\%, 1.6333/40\%, 2.8000/45\%, 3.9667/50\%}
  { \draw[tk] (0,\y) -- (-0.09,\y); \node[tick,anchor=east] at (-0.12,\y) {\l}; }
\node[font=\sffamily\footnotesize] at (3.50,-0.80) {Reasoning effort};
\node[font=\sffamily\footnotesize,rotate=90] at (-1.02,2.10) {Faithfulness};
\draw[oaiNavy,line width=1.3pt,line join=round,line cap=round]
  (0.70,0.9100) -- (2.10,2.3800) -- (3.50,3.4533) -- (4.90,3.5233) -- (6.30,3.8033);
\foreach \x/\y in {0.70/0.9100, 2.10/2.3800, 3.50/3.4533, 4.90/3.5233, 6.30/3.8033}
  \fill[oaiNavy] (\x,\y) circle (0.075);
\end{tikzpicture}%
\caption{gpt-5.6-sol's faithfulness with Lugbara input at each reasoning effort, against the English reference (95\% intervals about $\pm3$ points).}
\label{fig:effort}
\end{figure}

Figure~\ref{fig:effort} scores gpt-5.6-sol's Lugbara-input summaries at every reasoning effort it accepts against the English reference, since the judge cannot verify claims against a Lugbara article (Table~\ref{tab:judge-check}). More effort helps up to medium, from 0.37 without reasoning to 0.48, and little beyond it (0.49 at xhigh).

\section{Recovery Control on FLORES}
\label{app:flores}
FLORES-200 sentences come in fragments of 3.6 consecutive sentences on average from 562 known pages, so running the pipeline on the flat pool with the grouping withheld (GPT-5.5 composer, Sonnet 4.6 gate, minimum four sentences, six rounds, seven for Persian) tests what it recovers: Table~\ref{tab:flores} shows that 92 to 94\% of the returned articles are exactly one original fragment, no article mixes unrelated pages, and when asked for articles that span sources the pipeline returned nothing in any of the three languages.

\begin{table}[H]
\centering
\small
\begin{tabular}{lrrr}
\toprule
 & \textbf{Persian} & \textbf{Dari} & \textbf{Tajik} \\
\midrule
Sentences in the pool & 2{,}009 & 2{,}009 & 2{,}009 \\
Source fragments (3.6 sentences each) & 562 & 562 & 562 \\
Fragments with at least 4 sentences & 275 & 275 & 275 \\
\midrule
Articles returned & 232 & 230 & 159 \\
Articles made of a single source fragment & 214 (92\%) & 214 (93\%) & 150 (94\%) \\
Of which the whole fragment & 156 & 160 & 122 \\
Of which in the original sentence order & 198 & 202 & 145 \\
Mean share of sentences from the main source & 0.973 & 0.975 & 0.982 \\
Distinct qualifying fragments recovered & 214 (78\%) & 214 (78\%) & 150 (55\%) \\
Cost (composer and gate) & \$9.4 & \$7.9 & \$10.4 \\
\midrule
Articles required to span sources & 0 & 0 & 0 \\
\bottomrule
\end{tabular}
\caption{With the FLORES grouping withheld, the pipeline reassembles the original fragments and composes nothing across sources.}
\label{tab:flores}
\end{table}

\section{Verbatim Copies of the Source Sentences}
\label{app:nonllm}

After normalising whitespace and punctuation, 18{,}894 of the 18{,}940 Arabic sentences (99.76\%) are exact copies of their pool entries, 43 differ only by a unified proper noun and 3 by meaning-preserving trims. On the transfer builds 98.9 to 99.9\% of the source sentences and 99.9 to 100\% of the English sentences are exact copies.

\section{Judge Validation Detail}
\label{app:judge}
\paragraph{Judge choice and cross-judging.}
The gate and the judge are always a different model family from the composer, since LLM judges can favor outputs of related models \citep{li2025preferenceleakage}, and they are frontier models rather than small ones, since prior work reports model-dependent scoring biases in LLM judges \citep{smalljudgebias2025,reliabilitywithoutvalidity2025} and ReIFE finds that the ranking of base models is largely stable across protocols while a protocol's effect depends on the base model \citep{liu2025reife}. We therefore spend the choice on the base model and keep the protocol standard: a FActScore-inspired decompose-and-verify judge \citep{min2023factscore} (Section~\ref{sec:method}). All LLM judging was validated against human annotations (Sections~\ref{sec:validation} and~\ref{sec:rq3}).

\paragraph{Judge evaluation on the released articles.}
Table~\ref{tab:judge-check} scores each claim's faithful-or-not verdict against the source article against its verdict against the English article, over all systems of each build. The two verdicts agree moderately on the Arabic dialects and Persian varieties, where the source reference still passes about half of the claims the English reference rejects except on Tajik and Acholi, and barely above chance on Lugbara and Ateso, where the two rankings also come apart.

\begin{table}[H]
\centering
\small
\begin{tabular}{lccccc}
\toprule
& \multicolumn{2}{c}{\textbf{Cohen's $\kappa$}} & \multicolumn{2}{c}{\textbf{Source-input claims}} & \\
\cmidrule(lr){2-3}\cmidrule(lr){4-5}
\textbf{Build} & source input & English input & false passed & true rejected & \textbf{Ranking $\tau$} \\
\midrule
Arabic dialects & 0.54 & 0.40 & 47\% & 4\% & 0.87 \\
Persian & 0.51 & 0.47 & 53\% & 3\% & 0.73 \\
Dari & 0.57 & 0.48 & 49\% & 2\% & 1.00 \\
Tajik & 0.79 & 0.45 & 21\% & 3\% & 1.00 \\
Lugbara & 0.21 & 0.06 & 58\% & 19\% & 0.29 \\
Acholi & 0.63 & 0.17 & 27\% & 9\% & 0.93 \\
Ateso & 0.18 & 0.05 & 58\% & 24\% & 0.29 \\
\bottomrule
\end{tabular}
\caption{Agreement of the judge's verdicts against the source and the English article on the same claims, with the English verdict as truth. False passed and true rejected are for source-input claims; $\tau$ compares the systems' rankings under the two references.}
\label{tab:judge-check}
\end{table}

\section{Automatic Anchors}
\label{app:anchors}
All anchors are open-weight models run on every judged article and summary: AlignScore-large and MiniCheck with the English article as premise, mDeBERTa-v3-XNLI with the best sentence of either article, and multilingual-e5 retrieval of the closest sentence of either article. Scored claim by claim against the judge's English-reference verdict (Table~\ref{tab:anchors-nli}), every anchor that uses the source article is weaker than its English counterpart, and MiniCheck on the English article reaches AUC 0.81 to 0.93.

\begin{table}[H]
\centering
\small
\begin{tabular}{@{}lccccc@{}}
\toprule
\textbf{Checker} & \textbf{Arabic} & \textbf{Persian} & \textbf{Dari} & \textbf{Tajik} & \textbf{Lugbara} \\
\midrule
\multicolumn{6}{@{}l}{\textit{Uses the source article}} \\
\quad Judge, source reference & 0.737 & 0.739 & 0.751 & 0.869 & 0.569 \\
\quad mDeBERTa, source sentences & 0.585 & 0.598 & 0.588 & 0.668 & 0.469 \\
\quad e5 evidence, source sentences & 0.549 & 0.553 & 0.567 & 0.619 & 0.462 \\
\midrule
\multicolumn{6}{@{}l}{\textit{Uses the English article}} \\
\quad MiniCheck & 0.813 & 0.826 & 0.858 & 0.916 & 0.929 \\
\quad mDeBERTa, English sentences & 0.673 & 0.681 & 0.706 & 0.762 & 0.785 \\
\quad e5 evidence, English sentences & 0.648 & 0.625 & 0.652 & 0.777 & 0.871 \\
\midrule
Claims & 132{,}932 & 16{,}013 & 16{,}140 & 10{,}639 & 20{,}647 \\
\bottomrule
\end{tabular}
\caption{AUC of each anchor against the judge's English-reference verdict, per claim. Every anchor that uses the source article is weaker than its English counterpart.}
\label{tab:anchors-nli}
\end{table}

AlignScore, MiniCheck and English-side e5 rank the systems' source-input summaries in the order of the English-reference judge in every build (Table~\ref{tab:anchors-e5}), including the Lugbara reversal of Figure~\ref{fig:rankflip}, which therefore does not rest on the Claude judge; no anchor that uses the source article exceeds $\tau=0.20$.

\begin{table}[H]
\centering
\small
\begin{tabular}{@{}lccccc@{}}
\toprule
\textbf{Ranks the systems by} & \textbf{Arabic} & \textbf{Persian} & \textbf{Dari} & \textbf{Tajik} & \textbf{Lugbara} \\
\midrule
Judge, source reference & 0.87 & 0.73 & 1.00 & 1.00 & 0.29 \\
AlignScore-large, English article & 0.60 & 0.73 & 1.00 & 0.73 & 0.71 \\
MiniCheck, English article & 0.73 & 1.00 & 0.60 & 0.87 & 0.93 \\
e5 evidence, English sentences & 0.73 & 0.47 & 0.60 & 0.87 & 0.79 \\
mDeBERTa, English sentences & $-$0.20 & 0.47 & $-$0.07 & 0.73 & 0.86 \\
mDeBERTa, source sentences & $-$0.33 & 0.07 & $-$0.20 & $-$0.20 & $-$0.29 \\
e5 evidence, source sentences & $-$0.33 & $-$0.60 & 0.07 & 0.20 & $-$0.50 \\
\bottomrule
\end{tabular}
\caption{Kendall $\tau$ between each anchor's system ranking and the judge's English-reference ranking. Only anchors that read the English article agree with it.}
\label{tab:anchors-e5}
\end{table}

\section{Error Types}
\label{app:errors}
The error type is the judge's own tag for every claim below supported (Appendix~\ref{app:taxonomy}), and Table~\ref{tab:errortags} gives each tag's share of all claims, judged against the English article. Of the 4{,}293 entity-tagged dialect-input claims, 16\% give a name spelled like one in the article (e.g.\ Fahd for Fahad), which bounds transliteration at about one in six, and 26\% give a name absent from the article.

\begin{table}[H]
\centering
\small
\setlength{\tabcolsep}{4pt}
\begin{tabular}{lccccc}
\toprule
\textbf{Error tag} & \textbf{Arabic} & \textbf{Persian} & \textbf{Dari} & \textbf{Tajik} & \textbf{Lugbara} \\
\midrule
entity & 0.062 / 0.003 & 0.064 / 0.003 & 0.089 / 0.002 & 0.091 / 0.003 & 0.071 / 0.008 \\
number & 0.022 / 0.003 & 0.009 / 0.004 & 0.008 / 0.004 & 0.013 / 0.005 & 0.007 / 0.003 \\
unsupported & 0.132 / 0.054 & 0.053 / 0.035 & 0.061 / 0.029 & 0.198 / 0.035 & 0.536 / 0.044 \\
contradiction & 0.028 / 0.011 & 0.011 / 0.005 & 0.013 / 0.006 & 0.024 / 0.005 & 0.012 / 0.006 \\
relation & 0.073 / 0.059 & 0.051 / 0.038 & 0.058 / 0.037 & 0.071 / 0.033 & 0.032 / 0.074 \\
overgeneralization & 0.031 / 0.023 & 0.047 / 0.050 & 0.048 / 0.049 & 0.049 / 0.055 & 0.065 / 0.034 \\
\bottomrule
\end{tabular}
\caption{Per-claim error rates with source-language / English input, judged against the English article. Entity errors rise in every language, and half of the Lugbara-input claims are unsupported.}
\label{tab:errortags}
\end{table}

\section{Full Scores}
\label{app:dialect}
Table~\ref{tab:2x2} gives the faithfulness of both summaries against both articles for every system and variety; Figure~\ref{fig:summ} plots the gap between its English-reference scores.

\begin{table}[H]
\centering
\scriptsize
\setlength{\tabcolsep}{2.5pt}
\begin{tabular}{llcc@{\hskip 5pt}cc@{\hskip 5pt}cc@{\hskip 7pt}cc@{\hskip 5pt}cc@{\hskip 5pt}cc@{\hskip 5pt}cc@{\hskip 7pt}cc@{\hskip 5pt}cc@{\hskip 5pt}cc}
\toprule
 & & \multicolumn{6}{c}{\textbf{Persian varieties}} & \multicolumn{8}{c}{\textbf{Arabic dialects}} & \multicolumn{6}{c}{\textbf{Nilo-Saharan}} \\
 \cmidrule(lr){3-8}\cmidrule(lr){9-16}\cmidrule(lr){17-22}
 & & \multicolumn{2}{c}{Persian} & \multicolumn{2}{c}{Dari} & \multicolumn{2}{c}{Tajik} & \multicolumn{2}{c}{Egyptian} & \multicolumn{2}{c}{Gulf} & \multicolumn{2}{c}{Levantine} & \multicolumn{2}{c}{Maghrebi} & \multicolumn{2}{c}{Lugbara} & \multicolumn{2}{c}{Acholi} & \multicolumn{2}{c}{Ateso} \\
\textbf{System} & $r\backslash g$ & src & En & src & En & src & En & src & En & src & En & src & En & src & En & src & En & src & En & src & En \\
\midrule
\multirow{2}{*}{GPT-5.5} & src & .92 & .89 & .90 & .89 & .91 & .89 & .92 & .87 & .91 & .85 & .90 & .86 & .89 & .83 & .73 & .47 & .86 & .74 & .72 & .62 \\
 & En & .89 & .92 & .89 & .93 & .89 & .92 & .89 & .93 & .86 & .92 & .86 & .93 & .84 & .92 & .59 & .90 & .80 & .90 & .66 & .90 \\
\addlinespace[1pt]
\multirow{2}{*}{Qwen3-30B-A3B} & src & .86 & .79 & .85 & .76 & .73 & .66 & .80 & .72 & .77 & .67 & .76 & .66 & .73 & .57 & .55 & .21 & .29 & .17 & .60 & .16 \\
 & En & .86 & .89 & .87 & .88 & .86 & .88 & .87 & .89 & .83 & .87 & .82 & .88 & .80 & .88 & .54 & .88 & .79 & .88 & .62 & .88 \\
\addlinespace[1pt]
\multirow{2}{*}{gpt-oss-20B} & src & .83 & .74 & .82 & .71 & .67 & .58 & .83 & .73 & .78 & .68 & .80 & .67 & .80 & .65 & .73 & .40 & .45 & .26 & .70 & .31 \\
 & En & .85 & .88 & .87 & .89 & .86 & .89 & .82 & .86 & .80 & .84 & .83 & .85 & .81 & .86 & .55 & .84 & .77 & .84 & .64 & .84 \\
\addlinespace[1pt]
\multirow{2}{*}{Qwen3-8B} & src & .84 & .75 & .80 & .71 & .68 & .61 & .75 & .66 & .74 & .61 & .72 & .61 & .70 & .54 & .62 & .29 & .51 & .33 & .60 & .29 \\
 & En & .87 & .88 & .87 & .88 & .85 & .89 & .85 & .86 & .81 & .85 & .84 & .86 & .81 & .86 & .58 & .86 & .77 & .86 & .61 & .86 \\
\addlinespace[1pt]
\multirow{2}{*}{Aya-23-8B} & src & .85 & .81 & .83 & .76 & .19 & .16 & .78 & .69 & .73 & .62 & .71 & .62 & .69 & .55 & .34 & .09 & .11 & .06 & .38 & .04 \\
 & En & .84 & .86 & .84 & .85 & .82 & .84 & .76 & .79 & .76 & .79 & .77 & .78 & .75 & .79 & .55 & .83 & .76 & .83 & .62 & .83 \\
\addlinespace[1pt]
\midrule
\multirow{2}{*}{Specialist} & src & .83 & .76 & .79 & .70 & .69 & .61 & .83 & .72 & .79 & .67 & .78 & .66 & .77 & .60 & .48 & .50 & .75 & .65 & .53 & .66 \\
 & En & .84 & .84 & .85 & .86 & .84 & .87 & .80 & .84 & .80 & .84 & .80 & .85 & .78 & .85 & .53 & .84 & .74 & .84 & .60 & .84 \\
\addlinespace[1pt]
\multirow{2}{*}{Sunflower-32B} & src & \multicolumn{14}{c}{} & .52 & .61 & .75 & .67 & .59 & .65 \\
 & En & \multicolumn{14}{c}{} & .54 & .86 & .77 & .86 & .60 & .86 \\
\bottomrule
\end{tabular}
\caption{Faithfulness of summaries from the source ($r{=}$src) or English ($r{=}$En) input, checked against the source ($g{=}$src) or English ($g{=}$En) article. Specialists: Dorna2-8B (Persian), Fanar-1-9B (Arabic) and Sunflower-Qwen3.5-9B (Nilo-Saharan, whose English-input entries repeat), plus Sunflower-32B on the Nilo-Saharan languages.}

\label{tab:2x2}
\end{table}

\section{Article Length}
\label{app:length}
The released articles have 7 to 28 sentences (mean 8.1) and 25 to 329 English words (mean 74); the minimum sentences per article and the cluster size are set when the pipeline is run, so longer articles cost yield rather than a change of method, and Table~\ref{tab:length} splits the 2{,}170 judged articles into quartiles of English length.

\begin{table}[H]
\centering
\footnotesize
\setlength{\tabcolsep}{3.5pt}
\begin{tabular}{lccccccccc}
\toprule
\textbf{Quartile} & \textbf{Words} & \textbf{Sent.} & GPT-5.5 & gpt-oss & Qwen3-30B & Fanar & Qwen3-8B & Aya & \textbf{Gap} \\
\midrule
Q1 (shortest) & 25--58 & 7.3 & 0.922 & 0.840 & 0.829 & 0.834 & 0.793 & 0.769 & 0.070 \\
Q2 & 59--68 & 7.7 & 0.912 & 0.822 & 0.828 & 0.816 & 0.789 & 0.753 & 0.076 \\
Q3 & 69--83 & 8.2 & 0.916 & 0.825 & 0.816 & 0.804 & 0.797 & 0.757 & 0.074 \\
Q4 (longest) & 84--329 & 9.3 & 0.902 & 0.795 & 0.800 & 0.801 & 0.790 & 0.738 & 0.067 \\
\bottomrule
\end{tabular}
\caption{Faithfulness by quartile of English article length; the faithfulness gap stays near seven points.}
\label{tab:length}
\end{table}

\section{Where the Error Taxonomy Comes From}
\label{app:taxonomy}
The six error tags adapt the hallucination taxonomy of AraHalluEval \citep{alansari2025arahallueval} to a claim-versus-single-source setting (Table~\ref{tab:taxonomy}).

\begin{table}[H]
\centering
\footnotesize
\begin{tabular}{@{}p{2.8cm}p{6.1cm}p{3.6cm}@{}}
\toprule
\textbf{Tag} & \textbf{Definition} & \textbf{AraHalluEval analogue} \\
\midrule
entity & wrong or invented person, place, organisation or named thing & Named Entity (direct) \\
number & wrong or invented quantity, date, amount or count & Value (direct) \\
unsupported & content absent from the source & Fabrication, partly Context (direct) \\
overgeneralization& broader or stronger than the source supports & Inference (close) \\
contradiction & asserts the opposite of the source & Context Inconsistency (close) \\
relation & right entities, wrong relationship & none (extension) \\
\bottomrule
\end{tabular}
\caption{Each error tag maps onto a category of the AraHalluEval taxonomy.}
\label{tab:taxonomy}
\end{table}

\section{Qualitative Examples}
\label{app:qualitative}

\paragraph{An example cluster (Lugbara).}
Clustering gathers sentences on one topic, from which the composer picks a single scene. One cluster of the first round (recomputed with the settings of Appendix~\ref{app:models}) holds 129 of the 24{,}946 Lugbara pairs, all about health in Uganda; its English side includes ``Uganda is safe and has no Ebola cases'', ``Entebbe International Airport was closed as a measure to curb the spread of coronavirus'', ``The Uganda blood bank encourages people to donate blood'' and ``Cancer remains a silent killer in Uganda; therefore, we should all be screened''. From it the composer selected the fourteen sentences about one scene, the second wave of coronavirus in Uganda, which became benchmark article lgg-0203, beginning ``Uganda has been hit by the second wave of coronavirus. There were more severe cases in the second wave of coronavirus in Uganda. How many coronavirus cases are currently in Uganda?''

\paragraph{A Lugbara article that reads like news.}
SALT's sentences were written to cover ``a range of topics of local relevance, such as agriculture, health and society'' \citep{akera2022salt}, so the coherence gate rates the Lugbara articles close to the FLORES fragments of real documents and well above the Arabic phrasebook articles (Table~\ref{tab:gate-register}). Benchmark article lgg-0024, rated 5 out of 5, is one of them; its English side in full:
\begin{quote}\small
Museveni and Kagame were once comrades. Wars in Uganda, Rwanda, and the Democratic Republic of Congo were commanded by Museveni and his lieutenant Kagame. Since becoming president, Kagame has sought to break away from being Museveni's lieutenant. The Rwanda leadership now views Museveni as an enemy. It is difficult for leadership in Kigali to trust Museveni's commitment to any agreement. The Daily Monitor newspaper in Kampala in December twenty eighteen quoted Kagame. Rwandan President Paul Kagame spoke strongly about people trying to destabilize Rwanda. Kagame promised to raise the cost on anybody who wants to destabilize our security. Kagame's remarks were a broad sweep. Others said he aimed at either Burundi or Uganda. According to analysts, Kagame's tone was a harbinger of doom for the upcoming summit. The meeting was to follow up on an earlier meeting in Kigali. The meeting was aimed at resolving the stand-off with Uganda. He added that Uganda was determined to keep Rwanda on the negotiating table. Oryem says Uganda is willing to give talks another chance irrespective of changes in dates. Henry Okello Oryem is Uganda's minister of state for international affairs. Rwanda's minister of state wrote a letter to the bloc's secretary-general. State Minister for East African Community told about the false accusations made by Burundi. Uganda's Ministry of Foreign Affairs delivered a diplomatic protest in November. Uganda's Ministry of Foreign Affairs delivered a diplomatic protest. President Museveni and Paul Kagame have blocked open public discussion on the issues.
\end{quote}
These ratings come from the gate, not from native speakers; the human study covers the Arabic dialects only.

\begin{table}[H]
\centering
\small
\begin{tabular}{lccccc}
\toprule
\textbf{Build} & \textbf{Coherence} & \textbf{Rated 5} & \begin{tabular}[b]{@{}c@{}}\textbf{Reads like}\\\textbf{a document}\end{tabular} & \textbf{Summarizability} & \textbf{Sent./art.} \\
\midrule
Arabic dialects & 4.07 & 7\% & 2.70 & 2.93 & 8.1 \\
Lugbara & 4.71 & 71\% & 3.58 & 4.35 & 14.9 \\
Persian (FLORES fragments) & 4.89 & 89\% & 3.95 & 4.41 & 4.3 \\
Dari (FLORES fragments) & 4.91 & 91\% & 3.92 & 4.43 & 4.4 \\
Tajik (FLORES fragments) & 4.95 & 95\% & 4.01 & 4.52 & 4.4 \\
\bottomrule
\end{tabular}
\caption{The coherence gate's 1--5 ratings of every accepted article. Lugbara articles rate close to the FLORES fragments of real documents and well above the Arabic ones.}
\label{tab:gate-register}
\end{table}

\begin{figure}[H]
\centering
\resizebox{0.80\textwidth}{!}{%
\input{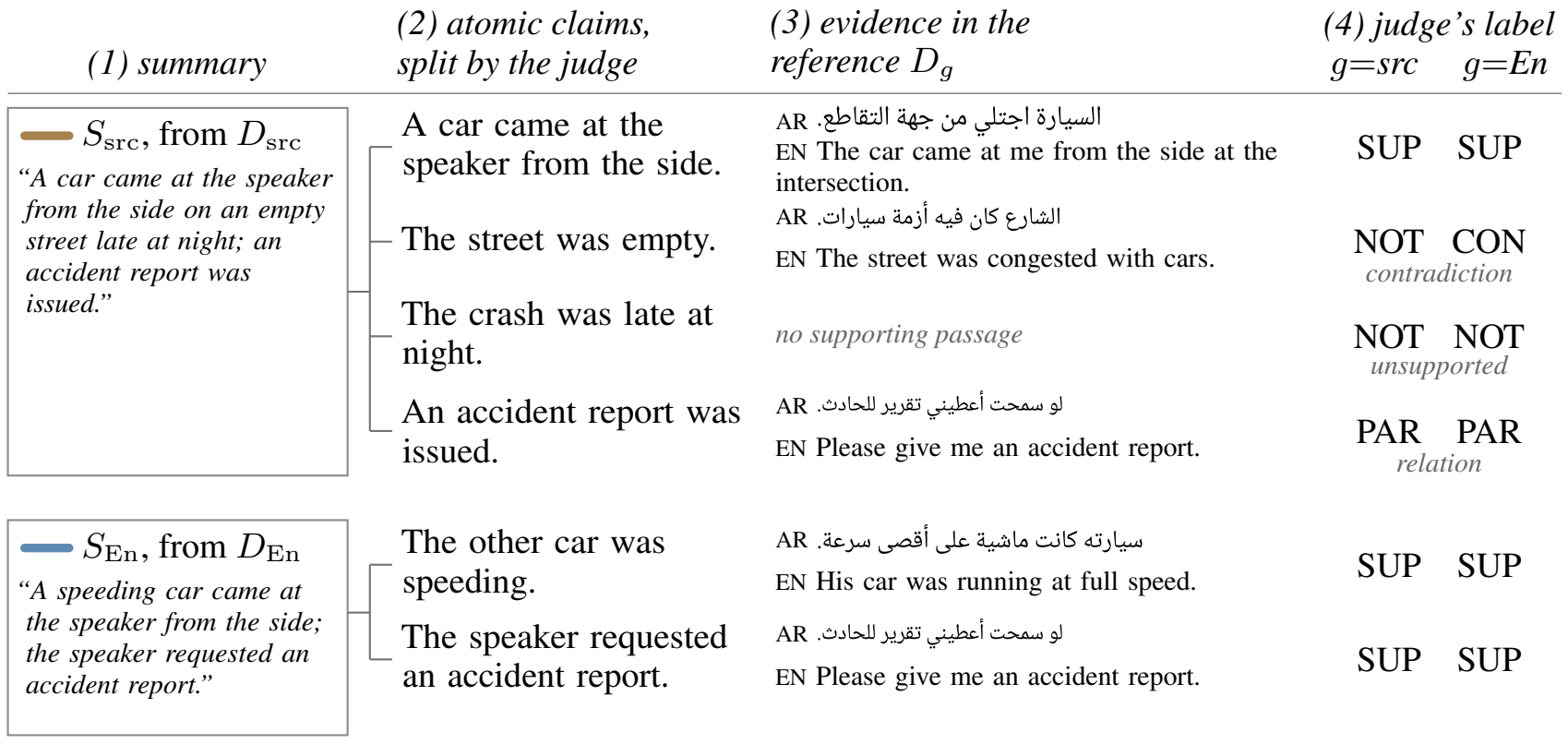}%
}
\caption{
Example of faithfulness evaluation, with Arabic as the low-resource language. Each of the two summaries, $S_{\mathrm{src}}$ from the low-resource input and $S_{\mathrm{En}}$ from the English input, is decomposed once into its atomic claims. Every claim is then checked against the low-resource reference $D_{\mathrm{src}}$ and against the English reference $D_{\mathrm{En}}$, receiving a label, the evidence span the label rests on, and an error tag (gray) for anything below supported: entity, number, unsupported, overgeneralization, contradiction or relation (Appendix~\ref{app:taxonomy}). Against the English reference the claims of $S_{\mathrm{src}}$ receive one of each label; against the Arabic reference the judge misses that the street was congested, the kind of failure the judge evaluation isolates.
}
\label{fig:claims}
\end{figure}

\section{Translation Quality and Faithfulness}
\label{app:translation}
Each open-weight system also translated every source sentence of the judged articles into English (prompt in Appendix~\ref{app:prompts}, greedy decoding), scored with chrF against the human English and averaged per article (Table~\ref{tab:translation}). Over all 27 system-build pairs mean chrF and $F_{\mathrm{src},\mathrm{En}}$ correlate at $r=0.93$, and on Lugbara only the two Sunflower models translate at all (chrF 47 and 54 against 17 to 21).

\begin{table}[H]
\centering
\small
\setlength{\tabcolsep}{4pt}
\begin{tabular}{@{}lccccc@{}}
\toprule
\textbf{System} & \textbf{Arabic} & \textbf{Persian} & \textbf{Dari} & \textbf{Tajik} & \textbf{Lugbara} \\
\midrule
gpt-oss-20B & 49.7 / 0.68 & 61.7 / 0.74 & 60.4 / 0.71 & 50.1 / 0.58 & 18.7 / 0.40 \\
Qwen3-30B-A3B & 43.3 / 0.65 & 62.0 / 0.79 & 60.1 / 0.76 & 48.6 / 0.66 & 17.1 / 0.21 \\
Qwen3-8B & 41.0 / 0.60 & 60.4 / 0.75 & 59.2 / 0.71 & 48.6 / 0.61 & 18.2 / 0.29 \\
Aya-23-8B & 43.4 / 0.61 & 70.6 / 0.81 & 67.4 / 0.76 & 26.9 / 0.16 & 18.0 / 0.09 \\
Fanar-1-9B & 50.4 / 0.66 & -- & -- & -- & 20.6 / 0.10 \\
Dorna2-8B & -- & 60.1 / 0.76 & 58.4 / 0.70 & 48.7 / 0.61 & -- \\
Sunflower-9B & -- & -- & -- & -- & 47.1 / 0.50 \\
Sunflower-32B & -- & -- & -- & -- & 54.0 / 0.61 \\
\midrule
$r$ across systems & 0.86 & 0.78 & 0.65 & 0.98 & 0.83 \\
\bottomrule
\end{tabular}
\caption{Mean chrF of each system's sentence translations / its faithfulness with source input (-- where not run). Last row: Pearson $r$ across a build's systems.}
\label{tab:translation}
\end{table}

\section{Licensing}
\label{app:licensing}
Table~\ref{tab:licensing} lists every text source the release draws on. Model outputs, extracted claims, judge labels, metric scores, and the survey instrument are our own and ship with the benchmark.

\begin{table}[H]
\centering
\footnotesize
\begin{tabular}{@{}p{3.6cm}p{2.1cm}p{3.9cm}p{2.9cm}@{}}
\toprule
\textbf{Source} & \textbf{License} & \textbf{Redistributed} & \textbf{Modifications} \\
\midrule
AraBench public text release \citep{sajjad2020arabench}: the MADAR-derived sentence pairs of 25 cities \citep{bouamor2018madar} and the QAraC Qatari transcripts \citep{elmahdy2014qarac} & none stated (public text release); MADAR and QAraC credited & the sentences and English references of 2{,}320 of the 2{,}330 articles, with a city code and scene identifier per article & selection and ordering only; proper-noun unification on 110 sentences, applied to both sides \\
Multidialectal Arabic Parallel Corpus (MDC) \citep{bouamor2014mdc} & research only, no redistribution & nothing: the 10 articles containing its 36 sentences are withheld & -- \\
CrossSum, BBC Arabic articles \citep{bhattacharjee2023crosssum} & CC BY-NC-SA 4.0 & the 16 natural-news survey items (4 per dialect) and the article pairs of the clustering diagnostic & none; verbatim excerpts, non-commercial terms apply to these items \\
FLORES-200 \citep{nllb2022}, Persian, Dari and Tajik pools & CC BY-SA 4.0 & articles from the transfer runs & selection and ordering only \\
SALT \citep{akera2022salt}, Lugbara, Acholi and Ateso pairs with English & CC BY-SA 4.0 & the 240 Lugbara articles and their Acholi and Ateso versions & selection and ordering only \\
\bottomrule
\end{tabular}
\caption{Every text source except MDC is redistributed under its own terms, and no source sentence was rewritten beyond proper-noun unification.}
\label{tab:licensing}
\end{table}

\end{document}